%% file: main.tex
\documentclass[sigconf,nonacm,screen]{acmart}
\AtBeginDocument{%
  }

\input{main_preamble}

\begin{document}

\title{CVTSplat: Multi-view Reconstruction Using an Adaptive Centroidal Voronoi Tessellation}

\author{Briac Toussaint}
\orcid{0009-0005-4609-3454}
\correspondingauthor
\affiliation{%
  \institution{Max Planck Institute for Informatics}
  \city{Saarbrücken}
  \country{Germany}
}
\email{btoussai@mpi-inf.mpg.de}

\author{Diego Thomas}
\affiliation{%
  \institution{Kyushu University}
  \city{Fukuoka}
  \country{Japan}
}
\email{thomas@ait.kyushu-u.ac.jp}
\orcid{0000-0002-8525-7133}

\author{Jean-S\'ebastien Franco}
\affiliation{%
  \institution{Inria}
  \city{Grenoble}
  \country{France}
}
\email{jean-sebastien.franco@inria.fr}
\orcid{0000-0002-4721-1441}

\author{Edmond Boyer}
\affiliation{%
 \institution{Inria}
 \city{Grenoble}
 \country{France}
}
\email{edmond.boyer@inria.fr}
\orcid{0000-0002-1182-3729}

\author{Christian Theobalt}
\affiliation{%
  \institution{Max Planck Institute for Informatics}
  \city{Saarbrücken}
  \country{Germany}
}
\email{theobalt@mpi-inf.mpg.de}
\orcid{0000-0001-6104-6625}

\renewcommand{\shortauthors}{Toussaint et al.}

\begin{abstract}
Voronoi tessellations have recently emerged as an adaptive, space-partitioning discretization for shape and appearance reconstruction from multi-view images. However, the irregular cells they produce make the repeated reconstruction of the tessellation during optimization costly. We therefore explore adaptive centroidal Voronoi tessellations (CVTs), which extend standard CVTs with non-uniform cell sizes while preserving the regular, well-shaped cells that make them efficient to compute. In CVTSplat, cell size is modulated by the evolving signed distance field, concentrating sites near the zero level set while keeping cells nearly centroidal. This regularity lets us specialize the whole pipeline, with GPU algorithms for nearest-neighbor queries and approximate Voronoi construction, and a rasterizer tailored to small, nearly isotropic cells. Our Voronoi construction is 5–10× faster than general-purpose alternatives in this regime. On DTU and BlendedMVS, CVTSplat reaches geometric accuracy competitive with the best Gaussian-splatting surface methods while training in about 2.5 minutes on a single GPU.
\end{abstract}



% \includegraphics[width=\textwidth]{sampleteaser}

%\includegraphics[width=\textwidth]{sampleteaser}
%\caption{figure caption}
%\Description{figure description}

\maketitle

\input{paper}

\bibliographystyle{ACM-Reference-Format}
\bibliography{main}

\appendix

\input{main_suppmat}

\end{document}

%% file: main_preamble.tex
\usepackage{amsmath}
\usepackage{cancel}
\usepackage{graphicx}
\usepackage{wrapfig}
\usepackage{amsfonts}
\usepackage{algorithm}
\usepackage[noend]{algpseudocode}
\usepackage{placeins}
\usepackage{makecell}
\usepackage{booktabs}
\usepackage{multirow}
\usepackage[table]{xcolor}

\newcommand\ie{\textit{i.e.}}
\newcommand\eg{\textit{e.g.}}
\newcommand{\sg}[1]{\operatorname{sg}\!\left(#1\right)}

\definecolor{bestcolor}{RGB}{255,178,178}
\definecolor{secondcolor}{RGB}{255,204,153}
\definecolor{thirdcolor}{RGB}{255,244,153}

\newcommand{\best}[1]{\cellcolor{bestcolor}#1}
\newcommand{\secondbest}[1]{\cellcolor{secondcolor}#1}
\newcommand{\thirdbest}[1]{\cellcolor{thirdcolor}#1}

\newcommand{\besttext}[1]{{\setlength{\fboxsep}{0.75pt}\colorbox{bestcolor}{#1}}}
\newcommand{\secondtext}[1]{{\setlength{\fboxsep}{0.75pt}\colorbox{secondcolor}{#1}}}
\newcommand{\thirdtext}[1]{{\setlength{\fboxsep}{0.75pt}\colorbox{thirdcolor}{#1}}}

%% file: paper.tex
\begin{figure}
    \centering
    \includegraphics[width=1.0\linewidth]{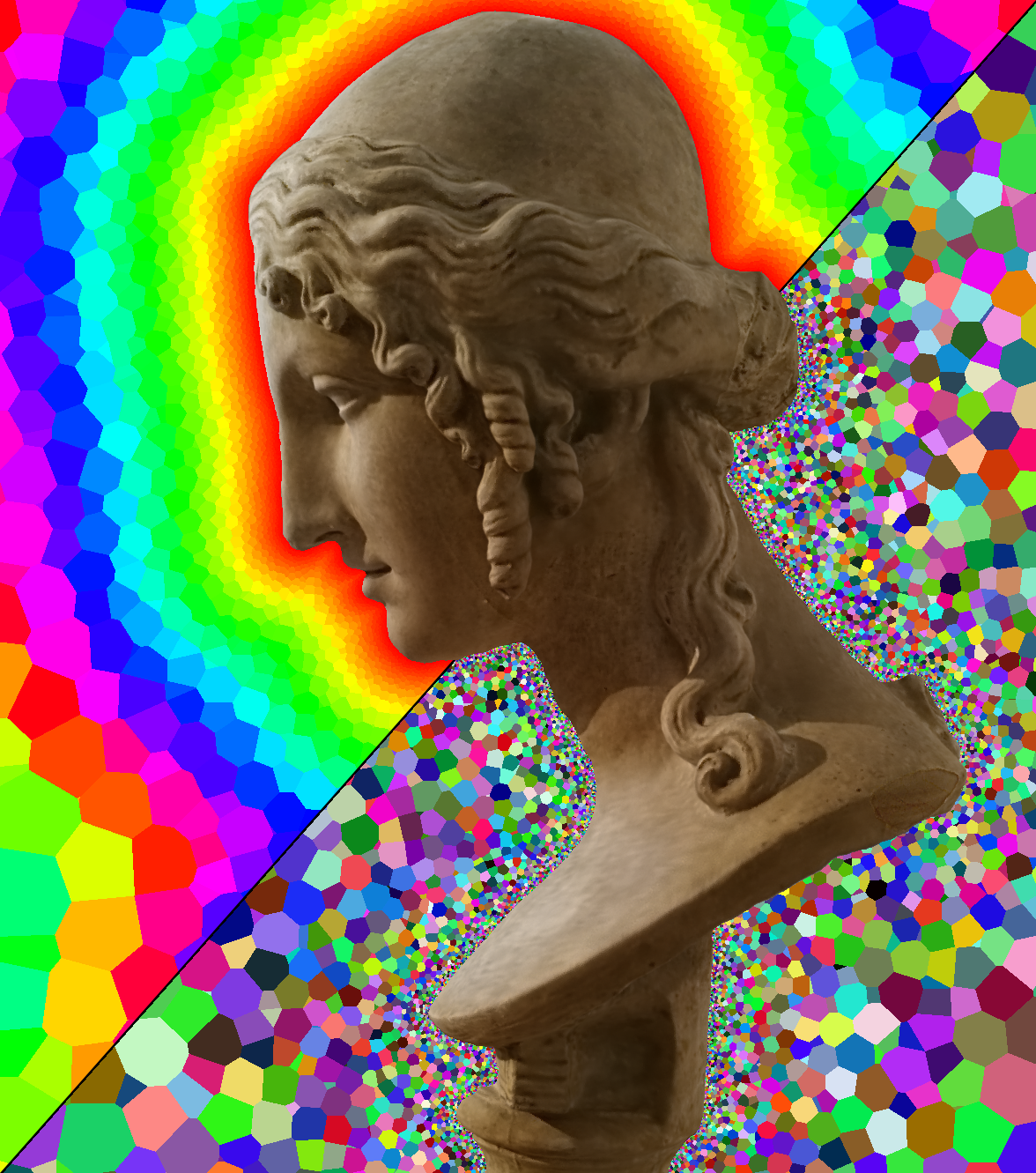}
    \caption{We represent the scene with Voronoi cells carrying trainable signed-distance and view-dependent appearance attributes, and directly rasterize them for differentiable volume rendering (middle). Our adaptive CVT combines regular cells with increased spatial resolution near the zero level set, concentrating representation capacity where it is most needed (bottom right). Top left: signed-distance values mapped to hue; bottom right: randomized cell colors illustrate the adaptive refinement near the surface.
    }
    \label{fig:teaser}
\end{figure}

\section{Introduction}

Multi-view reconstruction aims at recovering the geometry and appearance of a 3D scene from a collection of images acquired from different viewpoints. A central challenge is to design a representation that provides sufficient spatial resolution to capture fine surface details while remaining computationally efficient to optimize and render. Implicit surface methods coupled with differentiable rendering have demonstrated high reconstruction quality \cite{wang2021neus}, but their practical efficiency and compactness strongly depend on how the scene volume is discretized. Uniform discretizations, such as dense voxel grids, allocate the same capacity throughout space even though accurate geometry and appearance are primarily required near the reconstructed surface. Conversely, adaptive representations concentrate capacity in relevant regions, but introduce irregular spatial configurations that can make neighborhood queries, geometric operations, and rendering more expensive.

Voronoi-based representations provide an interesting direction for addressing this trade-off. Recent works have explored Voronoi and related cellular discretizations for neural rendering and reconstruction \cite{govindarajan2025radiant, govindarajan2026power, rech2026sdfoam, mai2026radiance, taveira2026vorotracing, weidenfeller2026adele}. However, these representations do not explicitly exploit one of the fundamental properties of Centroidal Voronoi Tessellations (CVTs): their tendency to produce regular, well-shaped cells. Such regularity is particularly attractive computationally, since irregular Voronoi cells may require visiting many neighboring sites when constructing the diagram \cite{ray2018meshless}. At the same time, a standard CVT tends toward a uniform distribution of sites, thereby losing the main benefit of an adaptive representation.

We therefore investigate adaptive centroidal Voronoi tessellations as a spatial representation for multi-view reconstruction. We modulate cell size using the implicit surface, yielding a near-centroidal tessellation with increased site density around the zero level set (fig. \ref{fig:teaser}). This combines the representational efficiency of an adaptive discretization with the computational benefits of regular Voronoi cells, while relying on a standard Voronoi diagram rather than a power diagram. Specifically, power diagrams offer another way to achieve adaptivity, but require an additional weight per site and incur a higher construction cost~\cite{govindarajan2026power, taveira2026scalable}.

Based on this representation, we introduce CVTSplat, a differentiable
multi-view reconstruction method in which each Voronoi site carries a
signed-distance value and compact appearance attributes decoded into
view-dependent radiance. The near-centroidal structure of the tessellation
lets us specialize every stage of the pipeline to this regime. Since
well-shaped cells have few and nearby neighbors, the Voronoi diagram can be
constructed from a small set of candidate sites. We deliberately accept an
approximate diagram in the rare cases where this set is incomplete, as the
resulting local overlaps are compatible with differentiable primitive-based
rendering~\cite{kerbl20233d}. Similarly, instead of ray tracing an exact
diagram~\cite{govindarajan2025radiant}, we directly rasterize the cells~\cite{govindarajan2026power} and
exploit their small, nearly isotropic shape to tailor the rasterizer.
Together, these choices make CVTSplat a regular yet surface-adaptive
discretization for efficient geometry and appearance reconstruction.

\begin{figure*}[h]
  \begin{center}
    \includegraphics[width=1.0\linewidth]{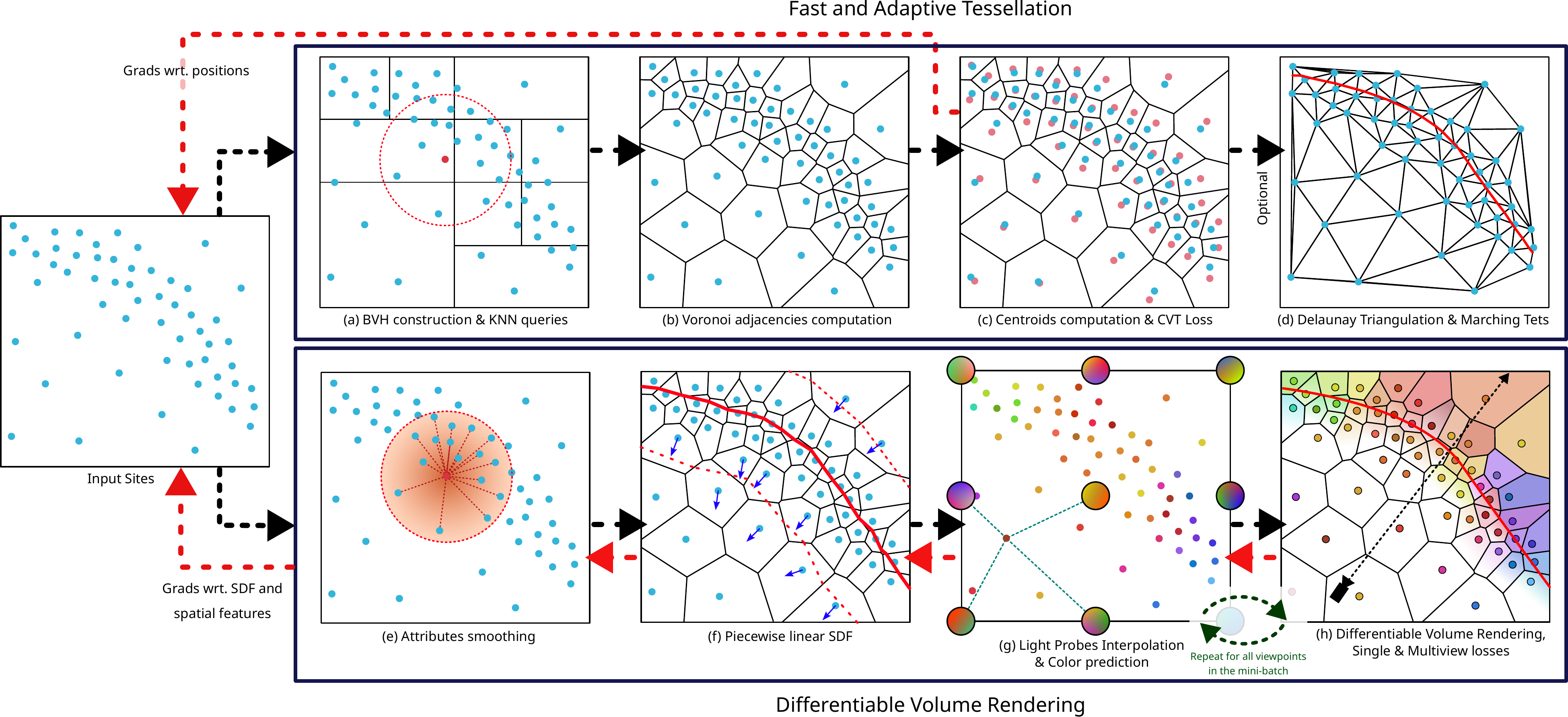}
  \end{center}
  \caption{Overview of our approach during optimization. Top row: Construction of the tessellation. (a) The sites are processed to obtain a list of their 32 nearest neighbors, (b) before building the Voronoi diagram as an adjacency graph. (c) The center of mass of each cell is then computed to regularize the distribution of the sites via the CVT loss. (d) The Voronoi diagram is optionally converted to a Delaunay triangulation to extract the zero-level set with marching tetrahedra. Bottom row: Differentiable volume rendering from posed images. (e) The sites' attributes (learnable SDF and features) are smoothed with a Gaussian kernel. (f) The gradient of the SDF is also computed to produce normal vectors, which define a piecewise linear SDF over the cells. Then, for all viewpoints in the mini-batch, (g) a per-site color is computed based on the camera's position, (h) before rasterizing the scene and comparing against the ground-truth image.}
  \label{fig:overview}
\end{figure*}

The complete optimization pipeline is described in figure~\ref{fig:overview}. It consists of two components: the adaptive tessellation controls the position of the sites, while the differentiable rendering branch optimizes the signed-distance and appearance attributes. The tessellation is guided by the evolving implicit surface to produce a surface-adaptive discretization. Our main contributions are as follows:

\begin{enumerate}
    \item We introduce an adaptive centroidal Voronoi discretization for multi-view reconstruction, combining surface adaptivity with the regularity of centroidal tessellations. Our complete reconstruction pipeline achieves geometric accuracy competitive  with state-of-the-art 3DGS-based methods \cite{chen2024pgsr, gomez2026blobs, Zhang2026GeometryGrounded} on DTU, while training in approximately 2.5 minutes on a single RTX 5090 and using 4 GB of VRAM.
        
    \item We exploit the regularity of the tessellation to make Voronoi-based reconstruction computationally efficient. We introduce specialized GPU algorithms for approximate Voronoi construction, repeated $K$-nearest-neighbor queries, and Delaunay triangulation, substantially accelerating these operations compared to general-purpose approaches \cite{ray2018meshless, basselin2021restricted, Cao2014gDel3D, taveira2026scalable, govindarajan2025radiant}. 

    \item We propose a rasterizer optimized for rendering small, isotropic primitives, further consolidating our framework.
\end{enumerate}
%Our CUDA implementation will be made publicly available upon publication for reproducibility.

\section{Related Work}

We review the literature most closely related to our approach along three directions: adaptive spatial discretizations for implicit 3D reconstruction (Sec.~\ref{sec:rel_discretizations}), explicit primitive-based representations (Sec.~\ref{sec:rel_explicit_primitives}), and efficient GPU construction of 3D Voronoi and Delaunay diagrams (Sec.~\ref{sec:rel_voronoi}).

\subsection{Adaptive Spatial Discretizations} \label{sec:rel_discretizations}

Multi-view 3D reconstruction based on differentiable rendering initially relied on continuous neural representations~\cite{mildenhall2021nerf, wang2021neus}, before increasingly adopting grid-based or hybrid spatial encodings to accelerate optimization and rendering~\cite{fridovich2022plenoxels, muller2022instant, chen2022tensorf, liu2020neural, toussaint2025probesdf}. The main objective of these representations is to allocate computation sparsely, by concentrating capacity around the reconstructed surface. However, grid-aligned discretizations typically follow a predefined spatial structure that either stays fixed during optimization or changes discretely across levels of detail.

Unstructured discretizations offer greater flexibility in adapting both the number and spatial extent of their elements. Recent approaches have explored Voronoi or Power cells~\cite{govindarajan2025radiant, govindarajan2026power, rech2026sdfoam, taveira2026vorotracing}, as well as tetrahedral representations~\cite{Thomas2025VortSDF, kulhanek2023tetra, mai2026radiance, weidenfeller2026adele}. 
% Briac: not sure about the next sentence, better to remove it. 
%Tetrahedral meshes have been applied to surface reconstruction~\cite{Thomas2025VortSDF, weidenfeller2026adele}, although their reconstruction accuracy remains below that of state-of-the-art grid-aligned approaches~\cite{li2026geosvr}.
These works motivate spatial representations that retain the flexibility of unstructured discretizations while achieving the accuracy and efficiency of highly optimized grid-based methods.

\subsection{Explicit Splatting Representations} \label{sec:rel_explicit_primitives}

3D Gaussian Splatting~\cite{kerbl20233d} has become a widely used representation for high-quality novel-view synthesis due to its adaptive distribution of primitives and efficient rasterization. However, recovering coherent geometry from an unstructured set of Gaussians is highly challenging~\cite{guedon2024sugar, Huang2DGS2024, yu2024gaussian, radl2025sof, chen2024pgsr, guedon2025milo, gomez2026blobs, Zhang2026GeometryGrounded}. Existing methods introduce additional constraints or regularization to suppress floaters and improve depth and normal consistency~\cite{ chen2024pgsr, Zhang2026GeometryGrounded}, or dedicated procedures for extracting closed surfaces~\cite{guedon2025milo, gomez2026blobs}.

Space-partitioning primitives such as tetrahedra and Voronoi cells provide an alternative splat support~\cite{govindarajan2025radiant, mai2026radiance} in which neighboring elements form a coherent spatial decomposition rather than an overlapping collection of primitives. This structure provides consistent spatial ordering, but maintaining it while the underlying sites move during optimization can introduce considerable computational overhead. Efficiently updating such adaptive representations therefore remains an important challenge.

\subsection{GPU Voronoi \& Delaunay Diagrams Construction} \label{sec:rel_voronoi}

A 3D Voronoi diagram can be represented through its adjacency graph, i.e., the neighboring relationships between generating sites. This graph can be recovered indirectly through the dual Delaunay triangulation~\cite{govindarajan2025radiant}, or computed directly from the Voronoi cells. GPU Delaunay algorithms enable substantial parallelism~\cite{Cao2014gDel3D, govindarajan2025radiant}, but generally require multiple rounds of point insertion, tetrahedron splitting, flipping, and repair, making frequent reconstruction of a dynamically evolving diagram costly.

GPU methods for direct Voronoi construction were initially developed for relatively uniform point distributions~\cite{ray2018meshless, basselin2021restricted}, providing large speedups over CPU implementations~\cite{CGAL2026, Levy2026Geogram}. More recently, Taveira et al.~\cite{taveira2026scalable} extended this family of approaches to non-uniform point distributions. Following the clipping formulation of Ray et al.~\cite{ray2018meshless}, Voronoi cells are constructed independently by successively clipping them against bisector planes induced by neighboring sites. Although cells can be processed in parallel, the clipping operations for each cell are sequential, and irregular cells may require significantly more operations. Leaving cells unregularized~\cite{govindarajan2025radiant} can thereby increase the total construction cost.

\section{Fast and Adaptive Spatial Data Structure} \label{sec:data_structure}

Our spatial representation consists of a set of sites $\{\mathbf{x}_i\}_{0\leq i<N}$ (\ie{} 3D points) that induce a Voronoi tessellation of the scene volume. Each site stores the geometric and appearance attributes used by the reconstruction, including a signed-distance value and view-dependent radiance features. The Voronoi cells provide the volumetric primitives used for rendering, while their dual Delaunay connectivity provides the local neighborhood structure required for mesh extraction. Because the site positions evolve throughout optimization, these structures must be updated repeatedly. A key property of our representation is that the sites remain close to a centroidal configuration, producing compact cells with a small number of nearby neighbors and enabling efficient GPU construction.

In the following, we first describe how we obtain this adaptive centroidal distribution (Sec. \ref{sec:adaptive_cvt}) and compute its centroids (Sec. \ref{sec:centroids}). We then present our fast Voronoi construction (Sec. \ref{sec:voronoi_diagram_construction}), its conversion to Delaunay tetrahedra (Sec. \ref{sec:delaunay_triangulation}), and the KNN kernel supporting these operations (Sec. \ref{sec:KNN}).

\subsection{Adaptive Centroidal Voronoi Tessellation} \label{sec:adaptive_cvt}

\begin{figure}[!htbp]
    \centering
    \includegraphics[width=1.0\linewidth]{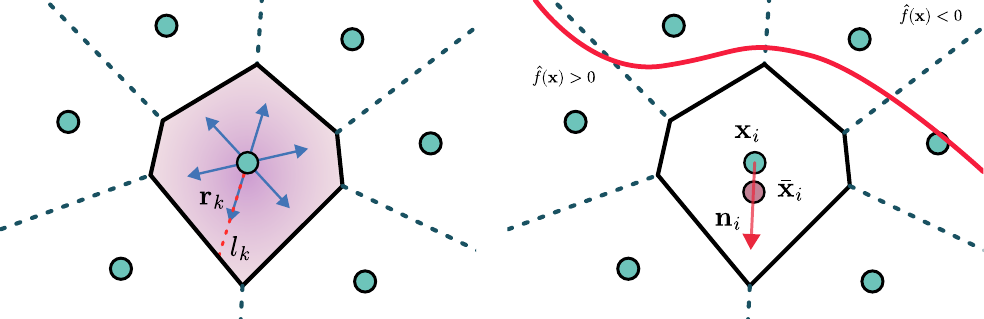}
    \caption{Left: The volume and centroid of the cell are computed by casting rays from the site's location. Right: Our biased CVT loss maintains the sites slightly closer to the implicit surface than the centroids $\bar{\mathbf{x}}$, resulting in stretched cells along the SDF gradient. }
    \label{fig:centroid}
\end{figure}

In a centroidal Voronoi tessellation (CVT), the sites $\mathbf{x}_i$ coincide with the center of mass $\bar{\mathbf{x}}_i$ of their cell (the centroids). It is not the case at first because the sites have arbitrary positions, but iteratively minimizing the CVT loss $\mathcal{L}_{\text{CVT}}=\sum ||\mathbf{x}_i - \bar{\mathbf{x}}_i||^2$ brings the tessellation closer to being centroidal, with regular cells \cite{lloyd1982least}. Since we need an adaptive distribution rather than a uniform one, we propose to bias this loss using the normal vector of the SDF (see figure \ref{fig:centroid}). Our biased CVT loss is as follows:
\begin{align}
    \mathcal{L}_{\text{CVT}} &= \sum_i \frac{1}{\sg{ r_i }} ||\mathbf{x}_i - \sg{\bar{\mathbf{x}}_i - \beta_i r_i \mathbf{n}_i}||_2^2 . \label{eq:CVTLoss}
\end{align}
$\sg{\cdot}$ is the stop-gradient operator and $r_i$ is the radius of the cell. Adaptivity may also be achieved using a spatially-varying mass density when computing cell moments~\cite{Du1999CentroidalVT} but our approach is more direct. Intuitively, the centroid $\bar{\mathbf{x}}_i$ is offset by a fraction $\beta_i$ of the cell radius along the SDF gradient $\mathbf{n}_i$. We divide by $r_i$ to make the loss scale-invariant. $\beta_i$ directly depends on the SDF $\hat{f}_i$:
\begin{align}
    \beta_i &= a \cdot \tanh(\frac{\hat{f}_i}{2b}) .
\end{align}
$a, b > 0$ control the magnitude of the shift. Using $a=0$ gives the standard centroidal tessellation. In our experiments, we set $a=0.1$ and $b=0.001$ which corresponds to shifting the centroid by $10\%$ of the cell radius. This results in smaller cells near the zero level-set and larger cells further away, with cell radius growing approximately linearly with SDF (refer to the supplementary material for an empirical analysis).  
The size of the cells is not explicitly imposed, which lets them arrange themselves independently of their number and of the object's surface area.

\subsection{Computation of the Centroids} \label{sec:centroids}

The CVT loss depends on the centroid $\bar{\mathbf{x}}_i$, itself depending on the volume $\mathcal{V}_i$. They can be computed using a tetrahedral decomposition of the cell \cite{ray2018meshless} or its polyhedron representation \cite{mirtich1996fast}. We propose a simpler alternative, although approximate, based on Riemann sums. Evaluating these sums only requires knowledge of the bisector planes, directly available from the Voronoi adjacency graph. Specifically, we sample $n=64$ rays $\mathbf{r}_k \in S^2$ from a Fibonacci sphere lattice, and measure the distance $l_k$ from the site $\mathbf{x}_i$ to the nearest bisector plane along $\mathbf{r}_k$ (see figure \ref{fig:centroid}, left). 
\begin{align}
    \mathcal{V}_i = \int_{\mathcal{C}_i} d \mathbf{x}= \int_{S^2} \int_0^{l(\mathbf{r})} s^2dsd\sigma = \int_{S^2} \frac{l(\mathbf{r})^3}{3} d\sigma \approx \frac{4\pi}{3n} \sum_{k=0}^{n-1} l_k^3 .\label{eq:cell_volume}
\end{align}
The volume integral over cell $\mathcal{C}_i$ is decomposed into a 1D integral and an integral over the unit sphere $S^2$. $l(\mathbf{r})$ is the distance from the site to the nearest clipping plane along $\mathbf{r}$ and $d\sigma$ is an infinitesimal surface element on $S^2$. Because the rays are approximately uniformly spread, their weight in the Riemann sum is $\frac{4\pi}{n}$.
A similar computation gives the centroid $\bar{\mathbf{x}}_i$:
\begin{align}
    \bar{\mathbf{x}}_i - \mathbf{x}_i&= \frac{1}{\mathcal{V}_i} \int_{\mathcal{C}_i} \mathbf{x} - \mathbf{x}_i d \mathbf{x} = \frac{1}{\mathcal{V}_i} \int_{S^2} \int_0^{l(\mathbf{r})} \mathbf{r} s \cdot s^2dsd\sigma \\ 
    &= \frac{1}{\mathcal{V}_i} \int_{S^2} \mathbf{r} \frac{l(\mathbf{r})^4}{4} d\sigma \approx \frac{1}{\mathcal{V}_i} \frac{4\pi}{n} \sum_{k=0}^{n-1} \mathbf{r}_k \frac{l_k^4}{4} = \frac{3}{4} \frac{\sum \mathbf{r}_k l_k^4}{\sum l_k^3} .
    \label{eq:cell_centroid}
\end{align}
$\mathcal{V}_i$ and $\bar{\mathbf{x}}_i$ are re-evaluated at every training iteration. The accuracy of these quadrature formulas is assessed in the appendix. Our approach does not preclude using a more direct computation~\cite{ray2018meshless, mirtich1996fast} if greater precision is required, since Voronoi vertices can be recovered from the Delaunay triangulation (Sec. \ref{sec:delaunay_triangulation}).

\subsection{Voronoi Diagram Construction} \label{sec:voronoi_diagram_construction}

Following previous GPU approaches \cite{ray2018meshless, basselin2021restricted, taveira2026scalable}, we represent the Voronoi diagram as an adjacency graph. However, we propose a new algorithm designed specifically for near-centroidal distributions. We assume the sites are in general position as a simplification, this means there are no duplicates and five sites or more are not cospherical.

\begin{figure}[h]
    \centering
    \includegraphics[width=0.8\linewidth]{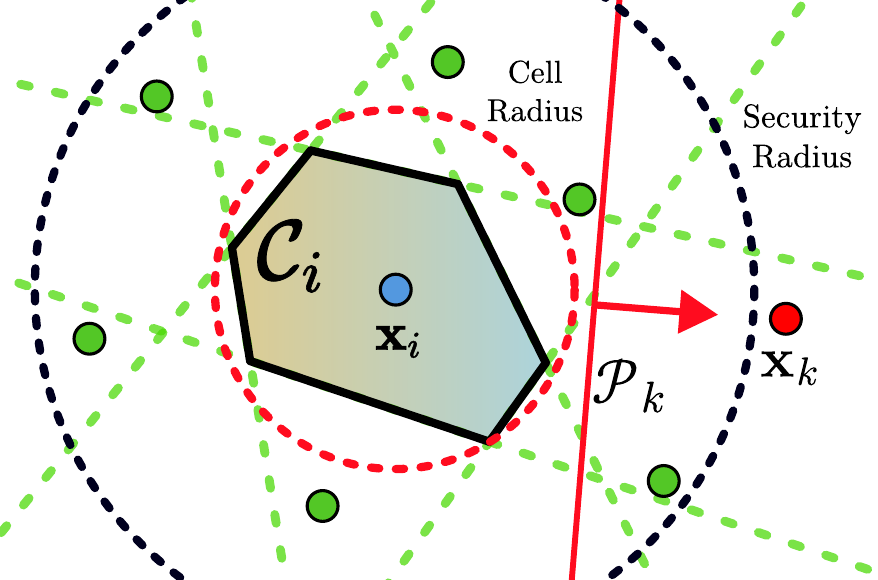}
    \caption{Voronoi cell in 2D. The bisector plane $\mathcal{P}_k$ is inactive since it does not participate in defining the cell $\mathcal{C}_i$. Any neighboring point beyond the security radius, defined as twice the cell radius, is guaranteed to not participate.}
    \label{fig:VoronoiDiagram2D}
\end{figure}

\paragraph{Cells as linear programs.} The Voronoi cell of the site $\mathbf{x}_i$ is:
\begin{align}
    \mathcal{C}_i = \{ \mathbf{x} \in \mathbb{R}^3, ||\mathbf{x} - \mathbf{x}_i||^2_2 \leq ||\mathbf{x} - \mathbf{x}_j||^2_2, \quad \forall j \neq i \} . \label{eq:voronoi_cell_def}
\end{align}
Let $\mathbf{x} = \mathbf{x}_i + \mathbf{y}$. The distance constraints in equation \ref{eq:voronoi_cell_def} can be reformulated geometrically as plane equations:
\begin{align}
    &||\mathbf{x} - \mathbf{x}_i||^2_2 \leq ||\mathbf{x} - \mathbf{x}_j||^2_2 \\
    \Leftrightarrow \quad &||\mathbf{y} ||^2_2 \leq ||\mathbf{x}_i + \mathbf{y} - \mathbf{x}_j||^2_2 \\
    \Leftrightarrow \quad &||\mathbf{y} ||^2_2 \leq ||\mathbf{x}_i - \mathbf{x}_j||^2_2 + 2 \mathbf{y} \cdot (\mathbf{x}_i - \mathbf{x}_j) + ||\mathbf{y}||_2^2 \\
    \Leftrightarrow \quad &\mathbf{y} \cdot (\mathbf{x}_j - \mathbf{x}_i) \leq \frac{1}{2} ||\mathbf{x}_j - \mathbf{x}_i||_2^2 . \label{eq:plane_constraint}
\end{align}
Therefore, the Voronoi cell is equivalently defined as the feasibility region of a linear program in $\mathbb{R}^3$:
\begin{align}
    \mathcal{C}_i = \{ \mathbf{x} \in \mathbb{R}^3, (\mathbf{x} - \mathbf{x}_i) \cdot (\mathbf{x}_j - \mathbf{x}_i) \leq \frac{1}{2} ||\mathbf{x}_j - \mathbf{x}_i||^2_2, \quad \forall j \neq i \} .\label{eq:voronoi_cell_def2}
\end{align}
The Voronoi cell $\mathcal{C}_i$ is bounded by a subset of the bisector planes between $\mathbf{x}_i$ and its neighboring sites. We call a neighbor $\mathbf{x}_j$ active if its bisector plane with $\mathbf{x}_i$ intersects $\mathcal{C}_i$, and the corresponding constraint  in (\ref{eq:voronoi_cell_def2}) contributes to the boundary of $\mathcal{C}_i$. Not all neighboring sites are necessarily active; for example, removing $\mathbf{x}_k$ does not change $\mathcal{C}_i$ in Fig. \ref{fig:VoronoiDiagram2D}. Therefore, identifying the set of active neighbors, or equivalently their bisector planes, is sufficient to define the Voronoi cell.

\paragraph*{Adjacency as 2D feasibility.} An active bisector plane must contain points of $\mathcal{C}_i$ since it corresponds to a facet of the convex polyhedron enclosing the cell. Conversely, inactive bisector planes do not contain any points of the Voronoi cell. This leads to a simple test:
\begin{align}
    &\text{The site } r \text{ is a Voronoi neighbor of site } i \\
    & \quad \Longleftrightarrow \text{$\mathcal{P}_r$ is active for cell $\mathcal{C}_i$} \\
    & \quad \Longleftrightarrow \mathcal{P}_r\cap\mathcal{C}_i\neq\emptyset
\end{align}
 
\begin{figure}[h]
    \centering
    \includegraphics[width=0.8\linewidth]{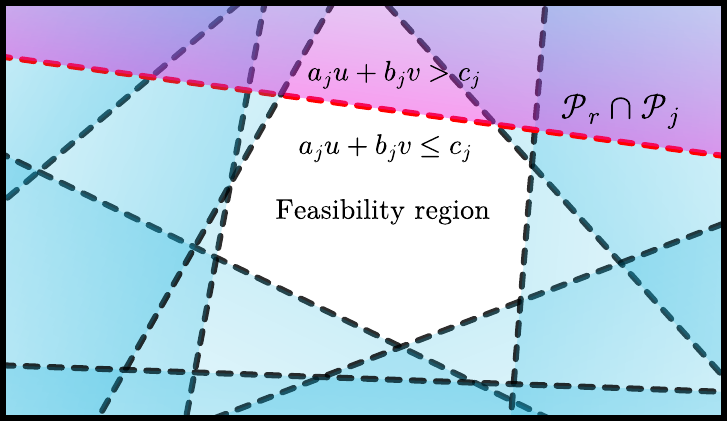}
    \caption{The 2D inequalities enclose the feasibility region, which corresponds to a facet of the polyhedron wrapping the cell. Each line is the intersection of the reference plane $\mathcal{P}_r$ with another bisector plane $\mathcal{P}_j$.}
    \label{fig:LinearProgram2D}
\end{figure}

We parametrize $\mathcal{P}_r$ as $\mathbf{x}=\tfrac12(\mathbf{x}_r+\mathbf{x}_i)+u\,\mathbf{e}_1+v\,\mathbf{e}_2$, where $(\mathbf{e}_1,\mathbf{e}_2, (\mathbf{x}_r-\mathbf{x}_i)/||\mathbf{x}_r-\mathbf{x}_i||)$ is an orthonormal basis~\cite{duff2017building}. Substituting the parametric equation for $\mathcal{P}_r$ into the remaining constraints of Eq.~\eqref{eq:voronoi_cell_def2} yields a 2D linear program (see Fig.~\ref{fig:LinearProgram2D}):
\begin{equation}
  a_j u + b_j v \le c_j,\qquad j\neq i,\; j\neq r,
  \label{eq:2dlp}
\end{equation}
and $\mathbf{x}_r$ is a neighbor if and only if this system is feasible.

%As a consequence we intersect $\mathcal{P}_r$ with the cell, and check whether the intersection is empty. In practice, we  translate the sites such that $\mathbf{x}_i = \mathbf{0}$, and rotate them such that $\mathcal{P}_r$ is above and parallel to the $xy$-plane. Such a rotation can be built by constructing an orthonormal basis from the unit vector $(\mathbf{x}_r - \mathbf{x}_i) / ||\mathbf{x}_r - \mathbf{x}_i||_2$ \cite{duff2017building}. In this reference frame, $\mathcal{P}_r = \{\mathbf{x}, z = \frac{1}{2} ||\mathbf{x}_r - \mathbf{x}_i||_2 \}$. Substituting the new plane equation for $\mathcal{P}_r$ into the other constraints (\ref{eq:voronoi_cell_def2}) gives a set of 2D inequalities, illustrated in figure \ref{fig:LinearProgram2D}:
%\begin{align}
%    a_jx+b_jy \leq c_j, \quad j \neq i \land j\neq r . \label{eq:seidel_2d_constraint}
%\end{align}

% This transformation gives a 2D linear program and the existence of a solution determines if site $r$ is a Voronoi neighbor of site $i$. Seidel's randomized algorithm \cite{seidel1991small} provides an efficient way of checking the existence of solutions (see \cite{lpalgorithmlecture} for an introduction). The constraints are not randomly shuffled in our case, we just take the sorted neighbors output by the KNN kernel (sec. \ref{sec:KNN}). 

\paragraph*{Feasibility test.}
We test feasibility with Seidel's incremental algorithm~\cite{seidel1991small,lpalgorithmlecture}, which maintains a feasible point and updates it only when a new constraint is violated, by solving a 1D problem along that constraint's boundary line. Instead of shuffling the constraints randomly, we process them in increasing distance order, as returned by the KNN kernel (Sec.~\ref{sec:KNN}). The nearest planes are the most likely to bound the cell, so infeasibility is usually detected early.

\paragraph*{GPU implementation.}
%We consider that only the 32 nearest neighbors are likely to define active bisector planes, so there are 32 threads dedicated for building each cell. The 2D constraints are evaluated on-the-fly and do not require per-thread storage. There are 37 constraints in total: 6 are provided by the scene bounding box and 31 by the neighbors of site $i$, different from site $r$. At the end, each thread outputs a single bit classifying whether the neighbor is Voronoi-adjacent.
We consider that only the 32 nearest neighbors are likely to define active bisector planes, so there are 32 threads dedicated for building each cell. Each one tests a candidate plane $\mathcal{P}_r$ against the 37 remaining constraints (31 other neighbors and the 6 planes of the scene bounding box). The 2D constraints are built on the fly from the site positions so no per-thread storage is needed, which is a significant bottleneck in previous approaches \cite{ray2018meshless, basselin2021restricted} due to reduced GPU occupancy. At the end, each thread outputs a single bit classifying whether the neighbor is Voronoi-adjacent.

\paragraph*{Approximation.} 
The assumption that 32 neighbors are enough generally holds in the case of our biased CVT. To further reduce the number of missing adjacencies, we center the KNN queries at the previously computed cell centroids rather than at the sites themselves. This substantially improves the completeness of the resulting Voronoi diagrams, as demonstrated in the supplementary material.
More neighbors could be included to handle more irregular cells, though with a performance loss. If a true neighbor beyond the $K$ nearest sites is missed, the cell will be enlarged and will overlap its neighbor. A missing constraint can also enlarge the feasible region of Eq.~\eqref{eq:2dlp}, so that a non-neighbor is wrongly marked as adjacent. Both errors are local and only affect cells near the few sites concerned.

\subsection{Delaunay Triangulation} \label{sec:delaunay_triangulation}

\begin{figure}[h]
    \centering
    \includegraphics[width=1.0\linewidth]{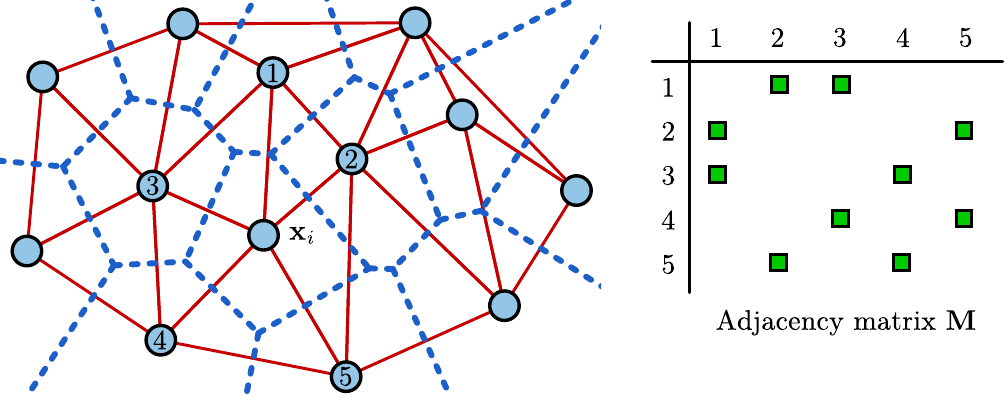}
    \caption{The adjacency matrix of the site $i$ encodes which of its neighbors are adjacent to each other.}
    \label{fig:AdjacencyMatrix}
\end{figure}

We convert the approximate Voronoi diagram into the dual Delaunay triangulation for mesh extraction, illustrated in figure \ref{fig:AdjacencyMatrix}. For conciseness, "Voronoi neighbor" will be shortened to "neighbor" in the following discussion.

The Delaunay tetrahedra are groups of 4 sites $(i, j, k, l)$ that are pairwise neighbors, and whose circumsphere does not contain any other site. Thus, we enumerate all possible 4-cliques, and then test whether their circumsphere is empty. Only ordered tuples $i<j<k<l$ are emitted to avoid duplicates.

We build a binary adjacency matrix M for each site $i$, that encodes the adjacency relation of its neighbors: $M_{j,k}$ is $1$ if sites $j$ and $k$ are neighbors. To form a quadruplet, we look for a neighbor $l$ of $i$ such that $M_{j,l}$ and $M_{k,l}$ are set to $1$ in M. 

We efficiently enumerate the quadruplets $(i, j, k, l)$ by leveraging the binary nature of M: because each site has 32 neighbors at most, the rows of M can be encoded as 32-bit integers. Thus, $j$ indices correspond to the direct neighbors of $i$, $k$ indices correspond to the bits set in row $M_j$, and $l$ indices correspond to the bits set in the binary AND of rows $M_j$ and $M_k$. The $\textit{\_\_ffs}$ CUDA intrinsic lets us iterate over the toggled bits of an integer. 

Not all quadruplets form Delaunay tetrahedra so we remove the ones containing other sites in their circumsphere. In practice, we only need to test the neighbors of $i$ against the circumsphere. A proof is available in the supplementary document.

Because our Voronoi diagram may be incomplete, our triangulation is not guaranteed to be exact for cells with missing clipping planes or in case of numerical ambiguities in the circumsphere check. This can result in a few holes or non-manifold edges in the extracted mesh out of millions of triangles. We quantify these defects in the supplementary material.

\subsection{Fast GPU KNN Search} \label{sec:KNN}

We propose a fast K-nearest neighbor kernel, tailored for $K=32$ and for repeated queries, to take advantage of our specific setting: the sites move by small amounts between rebuilds of the nearest-neighbors lists, which implies a high degree of temporal consistency that can be leveraged. Like Jakob et al. \cite{jakob2021optimizing}, we start by building an acceleration structure to avoid a naive $O(N^2)$ brute force check. We use a classic LBVH \cite{lauterbach2009fast} with up to 128 points per node. This gives a shallow hierarchy well-suited for querying a relatively large number of neighbors.

The nearest neighbors are gathered using ball-queries, \ie{} gathering all points within some distance of the query point, instead of the more common heap-based approach \cite{jakob2021optimizing, ray2018meshless}. %The initial search radius is taken as the diagonal of the BVH node containing the query point, divided by $\sqrt{3}$ to approximate its side length. 
The ball-query is restarted with a smaller or larger radius until the number of gathered points is between 32 and 64, using a binary search. They are subsequently sorted in increasing distance to the query point \cite{nvidia_cub_warpmergesort} and the first 32 are returned. Each query point has its own search radius which we save for later runs. The point cloud evolves slowly so the search radius is likely to be valid the next time the KNN routine is called, sparing repeated BVH traversals.

Heap-based approaches generally assign one thread per query point, and therefore need a per-thread traversal stack or priority queue, in addition to the heap already required to keep track of the K closest points. While the heap can be stored in registers \cite{jakob2021optimizing}, this reduces occupancy considerably for large values of K or spills the registers to global memory. 
%A traversal stack can be avoided if the nodes hold pointers to their parents, but requires pointer-chasing when walking up the tree. Traversing the tree incoherently also prevents coalescing memory accesses, both when visiting leaf nodes and when traversing the BVH. 

Our strategy addresses these issues: we assign one warp per query point, with all lanes visiting the BVH together. A single BVH traversal stack is needed for the whole warp, and the points can be loaded with coalesced memory accesses. 
The points within the search radius are compacted to a shared memory buffer sorted in parallel at the end, which is faster than iterative heap insertions. 

\section{Differentiable Volume Rendering} \label{sec:volume_rendering}

After gathering the nearest neighbors of all sites and computing the Voronoi diagram, we can proceed with rendering of the scene.
The volume rendering integral is typically evaluated with ray-marching \cite{mildenhall2021nerf, wang2021neus}, by querying the radiance and density fields at arbitrary locations along the rays. Directly adopting this formulation for our discretization is ill-suited: interpolating point neighborhoods at each ray sample is costly \cite{xu2022point}, while interpolating over the Voronoi cells generally requires a tetrahedral mesh~\cite{Thomas2025VortSDF, mai2026radiance}. Instead, we follow RadiantFoam \cite{govindarajan2025radiant} and use piecewise integration where each cell only contributes to the portion of the ray it intersects. RadiantFoam \cite{govindarajan2025radiant} and SDFoam \cite{rech2026sdfoam} use a constant density inside the cells which does not account for local surface orientation. In comparison, we define the SDF as a linear function in each cell to obtain nonlinearities once composed with NeuS \cite{wang2021neus}'s SDF-to-density conversion, and which crucially still gives a closed-form solution to the volume integral over the ray segments.

Three elements are necessary: (i) the SDF gradient at each site to define a piecewise SDF, (ii) per-primitive radiance and (iii) regularization terms to stabilize optimization. We reuse the nearest neighbors gathered during the Voronoi diagram construction (section \ref{sec:KNN}) to regularize the sites' attributes and to compute the SDF gradient.
For view-dependent appearance, we favor a neural implicit approach instead of per-site spherical harmonics to reduce the number of appearance parameters, as explained in section \ref{subsec:appearance_model}. Finally, section \ref{subsec:rasterization} covers our rasterization algorithm designed for small primitives.

\subsection{Attributes Smoothing} \label{subsec:geometric_representation}

Smoothing shape and appearance attributes is essential to stabilize optimization and recover sensible geometry in the case of discrete representations \cite{wu2022voxurf, toussaint2025probesdf}. Here, we leverage ideas from the Smoothed Particle Hydrodynamics (SPH) literature to define smoothing operators of arbitrary fields defined at the sites~\cite{price2012smoothed}. These operators are differentiable with respect to spatial coordinates, which in turn enables us to define the gradient of the SDF in a differentiable manner. Let $G(u)$ be the smoothing operator:
\begin{align}
    G(u)_i = \sum_{j \in \mathcal{N}(i)} u_j w_{ij} \ \  / \sum_{j \in \mathcal{N}(i)} w_{ij}. \label{eq:smoothing_kernel}
\end{align}
$u$ can be any scalar or vector field held by the sites, and $G(u)_i$ denotes the smoothed field computed at the position of site $i$. $\mathcal{N}(i)$ is the set of neighbors of $i$, plus $i$ itself. The smoothing weights are defined as follows:
\begin{align}
    w_{ij} &= \mathcal{V}_j \exp{\left(-\frac{||\mathbf{x}_i-\mathbf{x}_j||_2^2}{2\sigma^2 R_i^2} \right)} \\
    R_i &= \max_{j \in \mathcal{N}(i)}(||\mathbf{x}_i-\mathbf{x}_j||_2).
\end{align}
$\mathcal{V}_j$ is the volume of cell $j$ and $\sigma$ is chosen so that sites further away than the last neighbor would have small weights if they were included. Other smoothing schemes are possible \cite{koschier2020smoothed, sphtutorial} but we found this simple one to be effective.
The weights $w_{ij}$ take the cell volume into account so that $G$ approximates the convolution of the field $u$ by a Gaussian kernel:
\begin{align}
    (u(\mathbf{y})*e^{-\frac{||\mathbf{y}||^2}{2\sigma^2 R^2}})(\mathbf{x}) &= \int_{\mathbb{R}^3} u(\mathbf{y}) \exp{\left(-\frac{||\mathbf{x}-\mathbf{y}||^2}{2\sigma^2 R^2}\right)} d\mathbf{y} \\
    &\approx \sum_j u_j \exp{\left(-\frac{||\mathbf{x}-\mathbf{x}_j||^2}{2\sigma^2 R^2}\right)} \mathcal{V}_j. \label{eq:field_convolution}
\end{align}
Equation \ref{eq:field_convolution} is the Riemann sum of the integral discretized over all the sites, that we further divide by the sum of the weights to normalize $G$. Following \cite{wu2022voxurf, toussaint2025probesdf}, we apply our Gaussian filter $G$ on the raw SDF $f$ to obtain smoother values:
\begin{align}
    \hat{f} = G(f)
\end{align}
Next, we obtain the gradient vector of the smoothed SDF by computing partial derivatives of $G$ applied on $\hat{f}$:
\begin{align}
    \mathbf{n}_i &= \nabla_{\mathbf{x}_i} G(\hat{f})_i.
\end{align}

We are now able to define regularization objectives typically used in multi-view implicit surface reconstruction:
\begin{align}
    \mathcal{L}_\text{SDF} &= \sum_i \frac{(f_i - \hat{f}_i)^2}{\max(|\sg{\hat{f}_i}|, 0.01)} \\
    \mathcal{L}_\mathbf{n} &= \sum_i ||\frac{\mathbf{n}_i}{||\mathbf{n}_i||_2} - \sg{G(\frac{\mathbf{n}_i}{||\mathbf{n}_i||_2})}||_2^2 \label{eq:normal_smoothing} \\
    \mathcal{L}_\text{Eik} &= \sum_i (||\mathbf{n}_i||_2 - 1)^2
\end{align}

$\mathcal{L}_\text{SDF}$ keeps $f$ and $\hat{f}$ close to each other, $\mathcal{L}_\mathbf{n}$ is a smoothness term on the normalized gradient vectors and $\mathcal{L}_\text{Eik}$ is the classic eikonal regularization \cite{gropp2020implicit}. $\sg{\cdot}$ is the stop gradient operator added to simplify the backward pass. Furthermore, we do not back-propagate loss gradients through the positions of the sites, by treating them as constants in this formulation: their positions are entirely managed by the CVT loss defined in section~\ref{sec:adaptive_cvt}. 

So far, the SDF was only defined at the sites. We extend it to an entire Voronoi cell using an affine approximation:
\begin{align}
    \hat{f}(\mathbf{x}) = \hat{f}_i + (\mathbf{x} - \mathbf{x}_i) \cdot \mathbf{n}_i,  \quad \forall \mathbf{x} \in \mathcal{C}_i .  \label{eq:sdf_linear_approx}
\end{align}
When considering all the cells, the signed-distance field will be piecewise linear with small discontinuities allowed between cells. The affine approximations let us obtain a closed-form solution to the volume rendering integral, as described in section \ref{subsec:rasterization}, which may not be possible with typical interpolation schemes.

\subsection{Appearance Representation} \label{subsec:appearance_model}
\begin{figure}[h]
    \centering
    \includegraphics[width=1.0\linewidth]{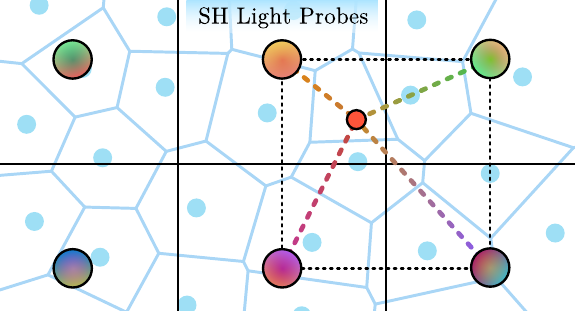}
    \caption{The light probes, parametrized with spherical harmonics, are interpolated to produce view-dependent features $\mathbf{F}^{a, \mathbf{v}}$ and $\mathbf{F}^{a, \mathbf{w}}$, based on the local view direction $\mathbf{v}$ and reflected vector $\mathbf{w}$.}
    \label{fig:light_probes}
\end{figure}
We assign a view-dependent color to each cell $\mathcal{C}_i$ to model the scene appearance. Like 3DGS \cite{kerbl20233d}, the color is constant inside a primitive, which gives a piecewise constant color field in our case. Our appearance parametrization takes inspiration from \cite{toussaint2025probesdf}. The main idea is to split view-independent components (\eg{} texture) from view-dependent components of the light field (\eg{} shading) to significantly reduce the number of parameters compared to a distinct spherical harmonic (SH) basis per primitive. 

The view-independent component is parametrized by a feature vector $\mathbf{F}_i^s \in \mathbb{R}^4$, called the spatial features, assigned to each site. The view-dependent component is modeled by a dense voxel grid of spherical harmonic (SH) coefficients called the light probes. They are trilinearly interpolated at the location of the sites to produce a per-primitive SH basis, subsequently evaluated in the direction of a unit vector to obtain angular features $\mathbf{F}_i^{a} \in \mathbb{R}^4$. More specifically, we use two sets of light probes: the first one is queried with the incident vector $\mathbf{v}$ and the other uses the reflected vector $\mathbf{w}$, which give $\mathbf{F}_i^{a, \mathbf{v}}$ and $\mathbf{F}_i^{a, \mathbf{w}}$ respectively. The site color $\mathbf{c}_i$ is obtained by decoding the spatial and angular features with a tiny MLP:
\begin{align}
\mathbf{c}_i = \text{MLP}(\mathbf{F}_i^s, \mathbf{F}_i^{a, \mathbf{v}}, \mathbf{F}_i^{a, \mathbf{w}}, (\mathbf{n}_i\cdot \mathbf{v}_i)^1, \cdots, (\mathbf{n}_i\cdot \mathbf{v}_i)^4) 
\end{align}

For regularization, we add a smoothness term on the spatial features, similar to the smoothness term on the normals:
\begin{align}
    \mathcal{L}_{\mathbf{F}^s} = \sum_i ||\mathbf{F}_i^s - \sg{G(\mathbf{F}_i^s)}||_2^2
\end{align}
The light probes are also regularized by applying a Gaussian filter $\mathcal{G}$ on the voxels and penalizing deviations between the smoothed and raw coefficients:
\begin{align}
    \mathcal{L}_\text{probes} = \sum ||\beta - \sg{\mathcal{G}(\beta)}||_2^2 \label{eq:probes_smoothing}
\end{align}
The sum is applied on all the voxels and SH basis coefficients $\beta$.

In practice, the two voxel grids have resolution $32^3$, with 16 SH coefficients per voxel and each SH coefficient in $\mathbb{R}^4$. The MLP has 16 neurons with 2 hidden layers and squareplus activations \cite{barron2021squareplus}. Because there are multiple sites located in the same voxel in general, the light probes represent a small proportion of all parameters (equivalent to $\approx1$ parameter per primitive in our experiments).

\subsection{Real-Time Rasterization} \label{subsec:rasterization}

The scene is ready to be rasterized into an image once every site has been assigned a color for the current camera position. We use NeuS' rendering equation \cite{wang2021neus} with reciprocal transmittance \cite{miller2024objects} to blend colors along a camera ray $\mathbf{o} + t \mathbf{r}, t >0$:
\begin{align}
    \mathbf{C} &= \int_0^{+\infty} \mathbf{c}(t) \sigma(t) T(0,t)dt \label{eq:volume_integral}\\
    T(0, t) &= e^{-\int_0^t \sigma(u)du} \\
    \sigma(t) &= s \Phi(-s\hat{f}(t)) |\mathbf{n}\cdot\mathbf{r}| \label{eq:sigma_neus_relu}
\end{align}
$\Phi$ is the sigmoid function and $s > 0$ is the sharpness parameter controlling the density decay outside of the implicit surface. In the equations above, $\mathbf{c}(t)=\mathbf{c}(\mathbf{o} + t \mathbf{r})$, $\sigma(t)=\sigma(\mathbf{o} + t \mathbf{r})$, etc.

\begin{figure}[h]
    \centering
    \includegraphics[width=1.0\linewidth]{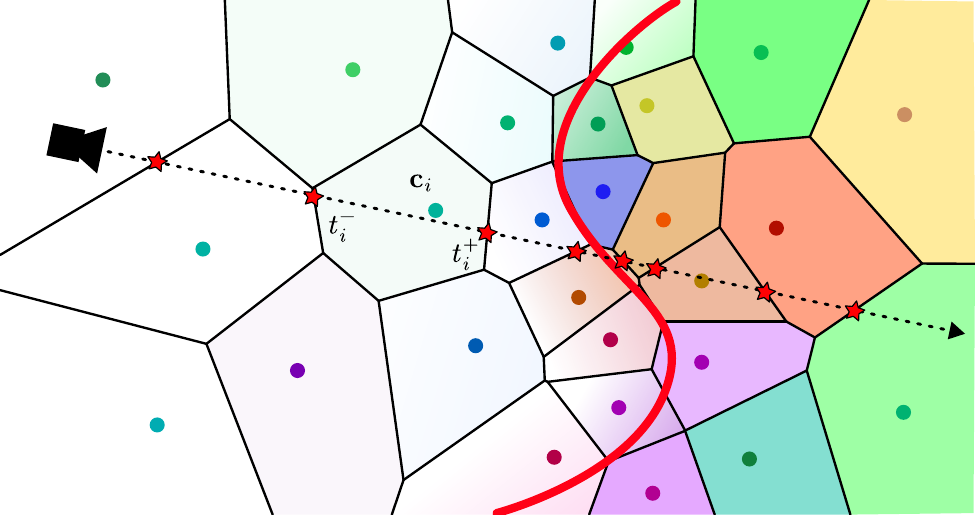}
    \caption{The color is constant but the density varies inside the cells. They are sorted by distance to the camera for pop-free rasterization.}
    \label{fig:rasterization}
\end{figure}

Because the color is constant inside a cell and the SDF is piecewise linear (see figure \ref{fig:rasterization}), and by further assuming the cells do not overlap which is typically the case, we obtain a closed-form solution for equation~\ref{eq:volume_integral}. Complete derivations are provided in the supplementary document, we give the final formula below:
\begin{align}
    \mathbf{C} &= \sum \mathbf{c}_i (1-\alpha_i) \prod_{j<i} \alpha_j \label{eq:blending}\\
    \alpha_i &= \frac{1+e^{-s\hat{f}(t_i^-)}}{1+e^{-s\hat{f} (t_i^+)}} \in [0,1] \label{eq:alpha}
\end{align}
We use the convention that $\alpha$ corresponds to the transparency of the cell. $\hat{f}(t_i^-)$ and $\hat{f}(t_i^+)$ are the piecewise SDF (equation \ref{eq:sdf_linear_approx}) evaluated at the entry and exit points of the ray in the cell respectively. The sum in equation \ref{eq:blending} is carried over the set of Voronoi cells intersected by the ray, ordered by increasing distance to the camera to guarantee geometrically-consistent blending (not depth, see \cite{govindarajan2026power} for a proof). Note that the fraction in eq. \ref{eq:alpha} should be inverted for back-facing cells so that $\alpha \leq 1$.  %Figure \ref{fig:rasterization} shows an illustration.

In practice, we first assign each cell to every $8\times8$ pixel tile intersected by the projection of its bounding sphere, sort them by tile id and by distance, and then evaluate equation \ref{eq:blending}. 

\subsubsection{Chunked Rasterizer} \label{subsec:chunked_rasterizer}

We noticed using one thread per ray results in poor GPU utilization whenever some rays intersect more cells than the average, which occurs for grazing rays that are parallel to the surface. In the worst case, a single thread can keep the GPU busy while the rest have terminated long before.
\begin{figure}
    \centering
    \includegraphics[width=1.0\linewidth]{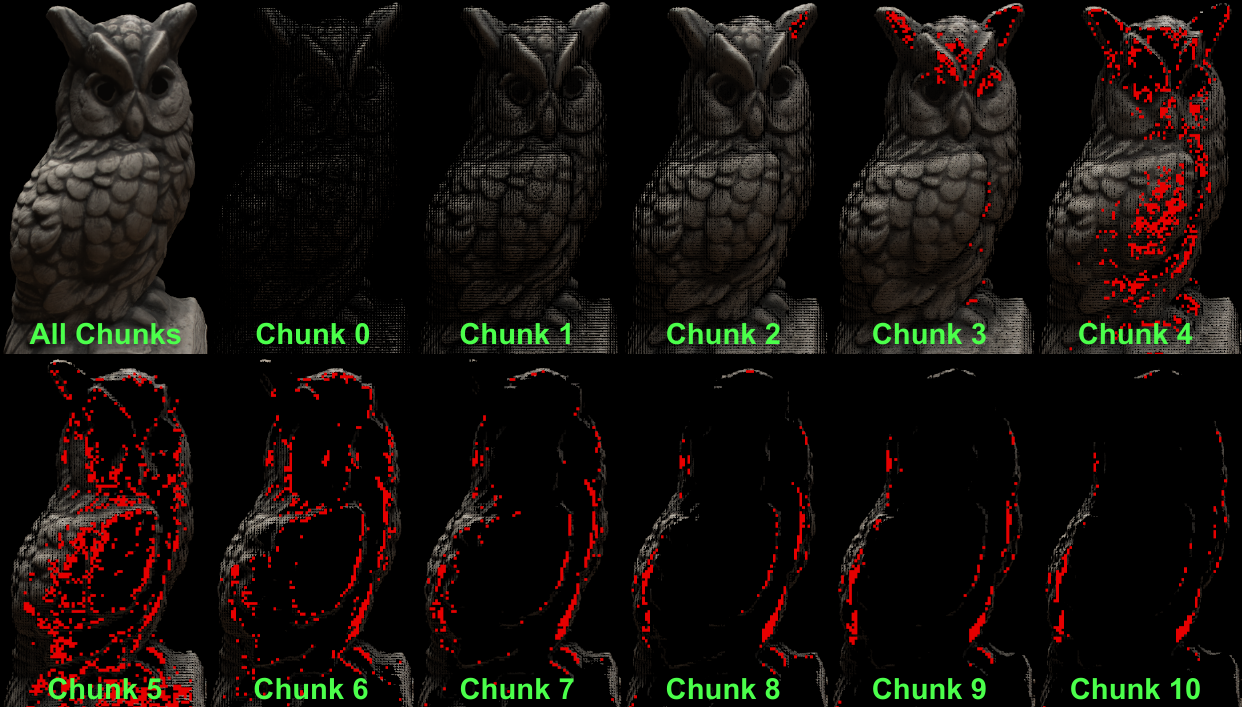}
    \caption{The pixel tiles are split into chunks of 32 consecutive cells. All chunks are rasterized in parallel but the ones marked in red have been skipped by detecting occlusions dynamically. The final result (top left) is the $\alpha$-blending of all the chunks in front-to-back order, evaluated in a subsequent pass.}
    \label{fig:ChunkRasterizer}
\end{figure}
Thus, we propose to split the sum in equation~\ref{eq:blending} into contiguous ranges of 32 cells that we call chunks, with the goal of increasing parallelism during rasterization.

As figure \ref{fig:ChunkRasterizer} shows, all chunks are scheduled to be processed in parallel, with chunks closer to the camera being handled first since we order them front-to-back. Naively processing the chunks all at once is not ideal because occluded cells would still be intersected and blended. In comparison, a sequential processing can exit as soon as the per-ray transmittance reaches zero. To retain this useful behavior, we propose a communication system based on per-pixel flags that are updated atomically by the chunks when they terminate their local rasterization task.

Each flag contains a transmittance accumulator $T_c$ and a distance counter $D_c$, updated jointly with an atomic compare-and-swap. $T_c$ corresponds to the multiplication of the partial transmittance values of the chunks having completed so far. The distance counter  $D_c$ contains the (depth) index of the furthest completed chunk. When a chunk starts, it reads the flags of its 8x8 pixel tile and decides if rasterization is needed based on the following table:
\begin{center}
\begin{tabular}{ |c|c|c| } 
 \hline
 Early-exit decision table & $T_c \geq \epsilon$ & $T_c$ < $\epsilon$ \\ 
 \hline
 chunk index < $D_c$ & Cannot skip & Cannot skip \\ 
 \hline
 chunk index > $D_c$ & Cannot skip & Can skip \\ 
 \hline
\end{tabular}
\end{center}

Rasterization is skipped if all rays in the chunk can skip. The accumulator $T_c$ is an upper bound of the true transmittance up to chunk $D_c$, valid for all chunks with depth index $i \geq D_c$. It is an upper bound since earlier chunks may not have updated $T_c$ when further ones read it. Therefore, a low value for $T_c$ guarantees that further chunks will not contribute to the final color. On the other hand, chunks closer than the distance counter have to run regardless of $T_c$. This scheme allows chunks pertaining to the same pixel tile to be processed fully in parallel if the GPU has enough thread blocks available, while hidden chunks will be opportunistically skipped. A distinct GPU kernel performs the final blending of each chunk's output color and transmittance in a later pass. 

The main cost in evaluating equation \ref{eq:blending} is the computation of the ray/cell intersection intervals $(t_i^-, t_i^+)$ due to execution divergence: each ray intersects a different subset of cells. To reduce this cost, we let threads work on different cells concurrently. The alternative is to iterate over the cells sequentially, and only the threads whose ray intersects the current cell can make progress while the others are inactive. For a given chunk, we first load the attributes of its 32 cells cooperatively in shared memory.

Then, each thread performs a ray/sphere intersection test \cite{haines2019precision} against the cells' bounding spheres, which produces a bit mask with one bit per cell. This part is sequential but relatively inexpensive. Next, each thread iterates over the bits set in its intersection mask %(we use the \textit{\_\_ffs} CUDA intrinsic to "peel" the mask bit-by-bit) 
and performs the more expensive ray/cell check against the corresponding cell, which need not be the same for all threads. The maximum number of bits set in the mask across threads determines the total number of iterations, often lower than $32$ since a ray is unlikely to intersect all cells' bounding spheres in the chunk. This optimization makes sense in our case because the cells are only a few pixels across and intersection tests are costly. Larger primitives would not benefit as much because more rays intersect them at once, and thus incur lower execution divergence.

The bisector planes are encoded as vectors $\mathbf{p}_i^k \in \mathbb{R}^3$, whose direction is the plane normal, oriented away from the site, and with length equal to the distance between the site and the plane. 
\begin{align}
t_i^- &= \max_{\substack{k\\ \mathbf{p}_i^k \cdot \mathbf{r}<0}}
\frac{
    \lVert \mathbf{p}_i^k\rVert_2^2
    -(\mathbf{o}-\mathbf{x}_i)\cdot\mathbf{p}_i^k
}{
    \mathbf{r}\cdot\mathbf{p}_i^k
},
\\
t_i^+ &= \min_{\substack{k\\ \mathbf{p}_i^k \cdot \mathbf{r}>0}}
\frac{
    \lVert \mathbf{p}_i^k\rVert_2^2
    -(\mathbf{o}-\mathbf{x}_i)\cdot\mathbf{p}_i^k
}{
    \mathbf{r}\cdot\mathbf{p}_i^k
}.
\end{align}
If $t_i^- \geq t_i^+$, this indicates the ray does not intersect the cell.

\section{Optimization and Implementation}

\subsection{Color remapping}

The rendered color $\mathbf{C}$ in equation \ref{eq:blending} is further composited with background pixel values sampled from a trainable envmap at infinity: $\mathbf{C}_2 = \mathbf{C} + \mathbf{C}_\text{env} T$, with $T=\prod \alpha_i$ the transmittance. We use an octahedral mapping to parametrize the envmap \cite{cigolle2014survey}, with a resolution of $1024^2$. Then, the color is tone-mapped to the $[0,1]$ interval: $\mathbf{C}_3=\mathbf{C}_2/(\mathbf{C}_2+1.0)$, before applying a trainable color correction curve: $\mathbf{C}_\text{final} = \gamma(\mathbf{C}_3)$. The color correction curve is a piecewise linear ramp mapping the $[0,1]$ interval to itself, distinct for each color channel and each camera \cite{toussaint2025probesdf}. 

\subsection{Image Losses}

Our data term is a weighted combination of a normalized L2 loss, a mask loss, a structural loss and a multi-view loss. We also add an L1 smoothing term on the depth, taken as the first intersection with the implicit surface.
\begin{align}
    \mathcal{L}_{\text{RGB}} &= \sum || \mathbf{C}_\text{final} - \mathbf{C}_{\text{gt}} ||^2 \frac{M}{\max{(\mathbf{C}_{\text{gt}}, 0.01)}} \\
    \mathcal{L}_{\text{Mask}} &= \sum (T - (1-M))^2 \\
    \mathcal{L}_{\text{DSSIM}} &= \sum (1-\text{SSIM}(\mathbf{C}_\text{final}, \mathbf{C}_{\text{gt}})) M \\
    \mathcal{L}_{\text{MV}} &= \sum (1-\text{ZNCC}) \\
    \mathcal{L}_{D} &= \sum |d - \sg{ \bar{d} }|
\end{align}
$\mathbf{C}_{\text{gt}}$ is the ground truth image, $M$ is the mask, $T$ is the transmittance, $\text{ZNCC}$ is the cross-correlation computed on pairs of nearby viewpoints, $d$ is the depth and $\bar{d}$ is the averaged depth of the neighboring pixels. The sums are applied to all the pixels of the cameras in the mini-batch. We use coarse-to-fine supervision from the ground truth images, with four levels of detail in total. Our complete optimization objective is a weighted combination of all the losses:
\begin{align}
    \mathcal{L} &= \lambda_{\text{RGB}} \mathcal{L}_{\text{RGB}} + \lambda_{\text{Mask}} \mathcal{L}_{\text{Mask}} + \lambda_{\text{DSSIM}} \mathcal{L}_{\text{DSSIM}} \notag\\
    &\quad + \lambda_{\text{MV}} \mathcal{L}_{\text{MV}} + \lambda_{D} \mathcal{L}_{D} + \lambda_{\text{CVT}} \mathcal{L}_{\text{CVT}}  \notag\\
    &\quad + \lambda_\text{SDF} \mathcal{L}_\text{SDF} + \lambda_\mathbf{n} \mathcal{L}_\mathbf{n} + \lambda_\text{Eik} \mathcal{L}_\text{Eik} \notag\\
    &\quad + \lambda_\text{probes} \mathcal{L}_\text{probes} + \lambda_{\mathbf{F}^s} \mathcal{L}_{\mathbf{F}^s} .
\end{align}

We propose to additionally weigh the image-loss gradients per-primitive during the backward pass based on their projected size. This re-weighting makes the training seamless when the resolution of the training images is switched to the next level of detail in the coarse-to-fine pyramid.
\begin{align}
    r_{\text{proj}} &= r f_x /z \\
    g_\text{mod} &= 1 / r_{\text{proj}}^2 \label{eq:grads_mod}
\end{align}
$r$ is the radius of the cell's bounding sphere, $f_x$ is the focal length and $z$ is the depth of the site for the current viewpoint. $g_\text{mod}$ is the per-primitive weight.

\subsection{Initialization and Densification}

Initially, we spread the sites in the scene bounding box based on a uniform distribution and apply 10 iterations of Lloyd's relaxation to uniformize the cells ($a=0$ in equation \ref{eq:CVTLoss}). The SDF is initialized from a sphere, while the rest of the learnable features are randomized based on a normal distribution.

We start with 200K sites, and increase their number every 10 iterations until we reach 4M sites by following a geometric progression. To place new sites, we pick a random viewpoint and render a depth map and a feature map. The feature map is the $\alpha$-blending of the sites' spatial features $\mathbf{F}^s$. A new site is placed at the depth sampled from a random valid pixel in the depth map, and its feature vector $\mathbf{F}^s$ is sampled from the same pixel in the feature map.

The nearest-neighbors list is rebuilt each time new points are added, while the Voronoi diagram is recomputed more frequently: at each iteration in the beginning, and every five iterations towards the end. We train for 10K iterations, with four training images per mini-batch. New points are added starting from the 400th iteration, until the last 400 iterations. For the last 400 iterations, we replace the planar-based ZNCC with a nonplanar version to avoid suppressing fine-scale details. More details regarding the multi-view loss and the weight scheduling are given in the supplementary material.

\section{Experimental validation}

All experiments in this section used a workstation equipped with an \textit{RTX 5090} GPU and an \textit{Intel Core Ultra 7 265K} with 20 cores to measure timings, with the exception of table \ref{tab:dtu_chamfer} where we report each method's timing and corresponding hardware configuration. \besttext{Best}, \secondtext{second} and \thirdtext{third} best results are highlighted in the tables.

\subsection{Multi-view Reconstruction}

\begin{figure}[!htbp]
    \centering
\includegraphics[width=1.0\linewidth]{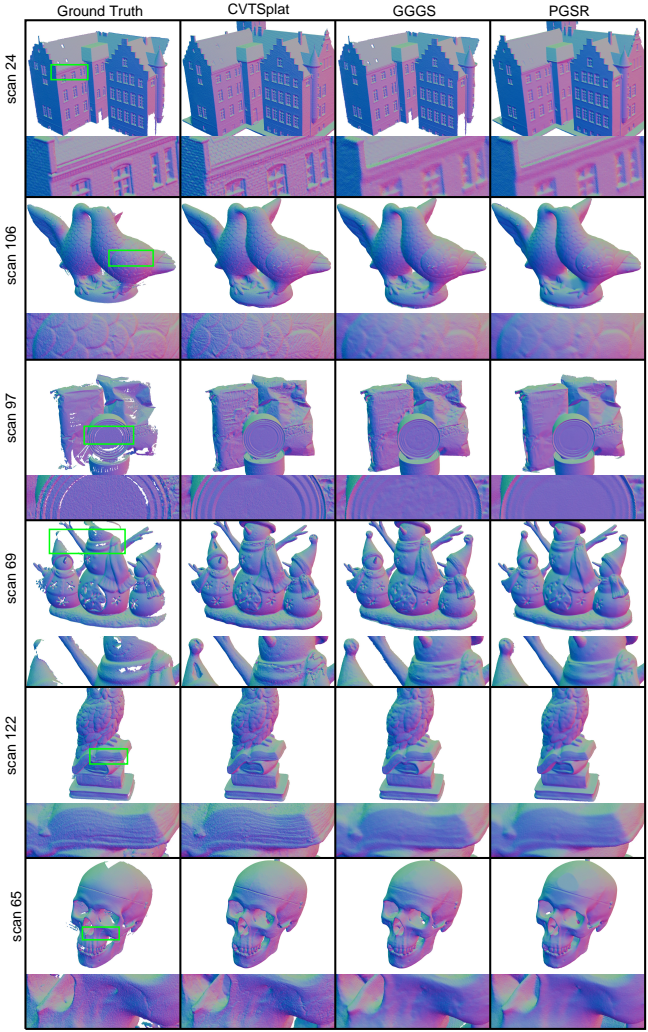}
    \caption{Qualitative comparison on DTU against GGGS \cite{Zhang2026GeometryGrounded} and PGSR \cite{chen2024pgsr}. The ground truth meshes were obtained by applying Poisson reconstruction \cite{kazhdan2013screened} on the ground truth point clouds. Our approach recovers finer details, especially visible on scans 24, 106 and 122.  }
    \label{fig:dtu_comparison}
\end{figure}

\textbf{Evaluation on DTU.} We evaluate the multi-view reconstruction accuracy of our approach on 15 scenes from the DTU dataset \cite{aanaes2016large}. Quantitative results are reported in table \ref{tab:dtu_chamfer} and qualitative results are shown in figure \ref{fig:dtu_comparison}. We achieve state-of-the-art geometric accuracy, tied with GGGS \cite{Zhang2026GeometryGrounded} and BlobsToSpokes \cite{gomez2026blobs}, while requiring a training time of approximately 2.5 minutes. Our approach reproduces high-frequency geometric details that GGGS and BlobsToSpokes miss, this is particularly noticeable in the close-ups on scans 24 and 106 shown at the top of fig. \ref{fig:dtu_comparison}. We hypothesize the qualitative improvement does not lead to quantitative improvement due to noise in the ground truth point clouds and down-sampling of the input reconstructions by the official evaluation script. 

For consistency with the 3DGS-like approaches \cite{Huang2DGS2024}, we discard vertices that project outside of the training masks as well as hidden vertices before Chamfer distance evaluation, since closed surfaces tend to be penalized under the DTU evaluation protocol. 

\begin{figure}[!htbp]
    \centering
\includegraphics[width=1.0\linewidth]{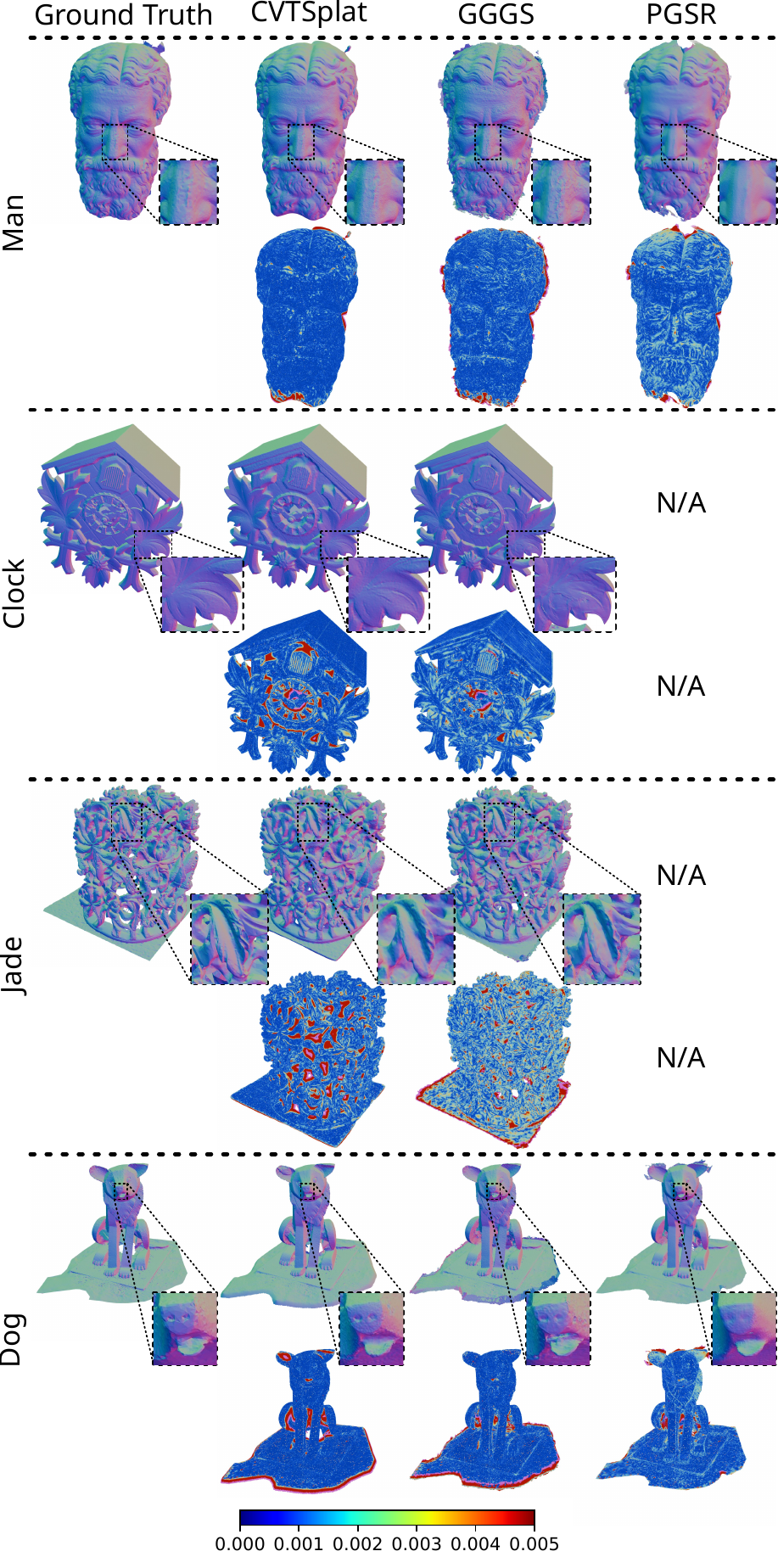}
    \caption{Qualitative comparison on BlendedMVS against GGGS \cite{Zhang2026GeometryGrounded} and PGSR \cite{chen2024pgsr}. Errors are mesured relative to the unit bounding box enclosing the objects. Our approach produces smoother geometry than GGGS (see close-ups on man and clock, heatmaps on jade), although some concavities are better reconstructed with GGGS (clock, jade, dog). PGSR fails on two examples and is less accurate overall. }
    \label{fig:bmvs_comparison}
\end{figure}

\noindent \textbf{Evaluation on BlendedMVS.} We also compare our approach on 8 scenes from the BlendedMVS \cite{yao2020blendedmvs} dataset in table \ref{tab:bmvs-chamfer}, with qualitative results presented in figure \ref{fig:bmvs_comparison}. We measure Chamfer distances as point-to-mesh distances in the normalized reference frame used for training, with the same evaluation protocol on this dataset as ProbeSDF~\cite{toussaint2025probesdf}. The images were cropped before running PGSR and GGGS according to the 2DGS protocol \cite{Huang2DGS2024}, since they require a projection center in the middle of the images. We achieve high geometric accuracy using the same hyperparameters as on DTU, with GGGS slightly ahead quantitatively but with noticeable noise (see figure \ref{fig:bmvs_comparison}). PGSR fails to produce a mesh in two instances due to ambiguous white backgrounds which caused the GPU to run out of memory during mesh extraction.

Additional evaluations of photometric accuracy are provided in the supplementary material.

\begin{table}[!htbp]
  \centering
  \caption{Chamfer distance on BlendedMVS with 4M sites.}
  \label{tab:bmvs-chamfer}
  \setlength{\tabcolsep}{4pt}
  \resizebox{\columnwidth}{!}{%
  \begin{tabular}{lccccccccc}
  \toprule
  Method & Mean & Dog & Bear & Clock & Durian & Man & Sculpture & Stone & Jade \\
  \midrule
  NeuS2 \cite{wang2023neus2} & 2.93 & 2.78 & 2.71 & 2.63 & 4.23 & 2.25 & 2.50 & 1.91 & 4.43 \\
  Voxurf \cite{wu2022voxurf} & 2.64 & 2.28 & 2.20 & 1.88 & 2.98 & 2.11 & 1.75 & 3.96 & 3.99 \\
  ProbeSDF \cite{toussaint2025probesdf} & 2.21 & 2.24 & 2.03 & \thirdbest{1.69} & 3.26 & 1.81 & \thirdbest{1.61} & \thirdbest{1.36} & \thirdbest{3.71} \\
  PGSR \cite{chen2024pgsr} & \thirdbest{1.56} & \thirdbest{1.21} & \thirdbest{1.20} & N/A & \thirdbest{2.15} & \thirdbest{1.39} & 1.81 & 1.58 & N/A \\
  GGGS \cite{Zhang2026GeometryGrounded} & \best{0.88} & \best{0.55} & \best{0.37} & \best{0.71} & \best{0.93} & \best{0.85} & \best{0.99} & \best{0.40} & \best{2.28} \\
  \midrule
  CVTSplat & \secondbest{1.13} & \secondbest{0.91} & \secondbest{0.49} & \secondbest{0.97} & \secondbest{1.44} & \secondbest{0.91} & \secondbest{1.21} & \secondbest{0.60} &
  \secondbest{2.53} \\
  \bottomrule
  \end{tabular}%
  }
\end{table}

\begin{table*}[t]
    \centering
    \caption{
        Chamfer distance on the DTU benchmark in millimeters.
        Lower is better. Best, second-best, and third-best results
        are highlighted.
    }
    \label{tab:dtu_chamfer}
    \setlength{\tabcolsep}{3.5pt}
    \renewcommand{\arraystretch}{1.05}

    \resizebox{\textwidth}{!}{%
    \begin{tabular}{cl*{15}{c}ccl}
        \toprule
        & \textbf{Method}
        & \textbf{24}
        & \textbf{37}
        & \textbf{40}
        & \textbf{55}
        & \textbf{63}
        & \textbf{65}
        & \textbf{69}
        & \textbf{83}
        & \textbf{97}
        & \textbf{105}
        & \textbf{106}
        & \textbf{110}
        & \textbf{114}
        & \textbf{118}
        & \textbf{122}
        & \textbf{Mean}
        & \textbf{Runtime}
        & \textbf{GPU} \\
        \midrule

        \multirow{8}{*}{\rotatebox{90}{Implicit}}

        % & NeRF \cite{mildenhall2021nerf}
        % & 1.90 & 1.60 & 1.85 & 0.58 & 2.28
        % & 1.27 & 1.47 & 1.67 & 2.05 & 1.07
        % & 0.88 & 2.53 & 1.06 & 1.15 & 0.96
        % & 1.49 & 24--48 h & V100 \\

        & VolSDF \cite{yariv2021volume}
        & 1.14 & 1.26 & 0.81 & 0.49 & 1.25
        & 0.70 & 0.72 & 1.29 & 1.18 & 0.70
        & 0.66 & 1.08 & 0.42 & 0.61 & 0.55
        & 0.86 & >12 h & RTX 2080 Ti \\

        & NeuS \cite{wang2021neus}
        & 1.00 & 1.37 & 0.93 & 0.43 & 1.10
        & 0.65 & 0.57 & 1.48 & 1.09 & 0.83
        & 0.52 & 1.20 & 0.35 & 0.49 & 0.54
        & 0.84 & 8 h & RTX 2080 Ti \\

        & Neuralangelo \cite{li2023neuralangelo}
        & 0.37 & 0.72 & 0.35 & 0.35 & 0.87
        & 0.54 & 0.53 & 1.29 & 0.97 & 0.73
        & 0.47 & 0.74 & \thirdbest{0.32} & 0.41 & 0.43
        & 0.61 & >128h & V100 \\

        & NeuS2 \cite{wang2023neus2}
        & 0.56 & 0.76 & 0.49 & 0.37 & 0.92
        & 0.71 & 0.76 & 1.22 & 1.08 & 0.63
        & 0.59 & 0.89 & 0.40 & 0.48 & 0.55
        & 0.70 & \secondbest{5 min} & RTX 3090 \\

        & Voxurf \cite{wu2022voxurf}
        & 0.91 & 0.73 & 0.45 & 0.34 & 0.99
        & 0.66 & 0.83 & 1.36 & 1.31 & 0.78
        & 0.53 & 1.12 & 0.37 & 0.53 & 0.51
        & 0.76 & 15 min & A100 \\

        & ProbeSDF \cite{toussaint2025probesdf}
        & 0.65 & 0.74 & 0.34 & 0.34 & 1.02
        & 0.71 & 0.62 & 1.34 & 0.94 & 0.70
        & 0.53 & 0.96 & 0.36 & 0.45 & 0.47
        & 0.68 & \best{2.5 min} & RTX 3090 \\

        & Geo-NeuS \cite{fu2022geoneus}
        & 0.38
        & 0.54
        & 0.34
        & 0.36
        & 0.80
        & \secondbest{0.45}
        & \best{0.41}
        & \secondbest{1.03}
        & 0.84
        & \thirdbest{0.55}
        & 0.46
        & \secondbest{0.47}
        & \best{0.29}
        & 0.36
        & 0.35
        & 0.51
        & $>12$ h
        & RTX 3090 \\

        & GeoSVR \cite{li2026geosvr}
        & \best{0.32}
        & \thirdbest{0.51}
        & \thirdbest{0.30}
        & 0.33
        & \best{0.71}
        & 0.48
        & \secondbest{0.42}
        & \secondbest{1.03}
        & \best{0.62}
        & 0.56
        & \secondbest{0.33}
        & \best{0.46}
        & \secondbest{0.30}
        & 0.34
        & \secondbest{0.32}
        & \best{0.47}
        & 50 min
        & RTX 3090 Ti \\

        \midrule

        \multirow{9}{*}{\rotatebox{90}{Explicit}}

        % & 3DGS \cite{kerbl20233d}
        % & 2.14 & 1.53 & 2.08 & 1.68 & 3.49
        % & 2.21 & 1.43 & 2.07 & 2.22 & 1.75
        % & 1.79 & 2.55 & 1.53 & 1.52 & 1.50
        % & 1.96 & 8 min & RTX 4090 \\

        & SuGaR \cite{guedon2024sugar}
        & 1.47 & 1.33 & 1.13 & 0.61 & 2.25
        & 1.71 & 1.15 & 1.63 & 1.62 & 1.07
        & 0.79 & 2.45 & 0.98 & 0.88 & 0.79
        & 1.33 & 1h & RTX 4090 \\

        & 2DGS \cite{Huang2DGS2024}
        & 0.48 & 0.91 & 0.39 & 0.39 & 1.01
        & 0.83 & 0.81 & 1.36 & 1.27 & 0.76
        & 0.70 & 1.40 & 0.40 & 0.76 & 0.52
        & 0.80 & 12 min & RTX 4090 \\

        & GOF \cite{yu2024gaussian}
        & 0.50 & 0.82 & 0.37 & 0.37 & 1.12
        & 0.74 & 0.73 & 1.18 & 1.29 & 0.68
        & 0.77 & 0.90 & 0.42 & 0.66 & 0.49
        & 0.74 & 50 min & RTX 4090 \\

        & RaDe-GS \cite{zhang2026rade}
        & 0.46 & 0.73 & 0.33 & 0.38 & 0.79
        & 0.75 & 0.76 & 1.19 & 1.22 & 0.62
        & 0.70 & 0.78 & 0.36 & 0.68 & 0.47
        & 0.68 & \thirdbest{8 min} & RTX 4090 \\

        & MILo \cite{guedon2025milo}
        & 0.43 & 0.74 & 0.34 & 0.37 & 0.80
        & 0.74 & 0.70 & 1.21 & 1.22 & 0.66
        & 0.62 & 0.80 & 0.37 & 0.76 & 0.48
        & 0.68 & 25 min & RTX 4090 \\

        & PGSR \cite{chen2024pgsr}
        & \secondbest{0.34}
        & 0.54
        & 0.44
        & 0.37
        & \thirdbest{0.78}
        & 0.57
        & 0.49
        & 1.06
        & \secondbest{0.63}
        & 0.59
        & 0.47
        & \thirdbest{0.50}
        & \secondbest{0.30}
        & 0.37
        & 0.34
        & 0.52
        & 30 min
        & RTX 4090 \\

        & GGGS \cite{Zhang2026GeometryGrounded}
        & 0.38
        & \secondbest{0.50}
        & \secondbest{0.27}
        & \secondbest{0.31}
        & 0.80
        & \best{0.43}
        & \secondbest{0.42}
        & \thirdbest{1.04}
        & \thirdbest{0.64}
        & \secondbest{0.52}
        & \best{0.31}
        & 0.56
        & \secondbest{0.30}
        & \best{0.31}
        & \thirdbest{0.33}
        & \best{0.47}
        & 15 min
        & RTX 4090 \\

        % & GGGS (30k iter)
        % & 0.37
        % & 0.51
        % & 0.27
        % & 0.31
        % & 0.81
        % & 0.43
        % & 0.42
        % & 1.05
        % & 0.64
        % & 0.52
        % & 0.32
        % & 0.58
        % & 0.30
        % & 0.31
        % & 0.33
        % & 0.48
        % & 25 min
        % & RTX 4090 \\

        & BlobsToSpokes \cite{gomez2026blobs}
        & 0.39
        & \secondbest{0.50}
        & \best{0.26}
        & \thirdbest{0.32}
        & 0.80
        & \thirdbest{0.46}
        & \thirdbest{0.44}
        & 1.05
        & 0.65
        & \secondbest{0.52}
        & \thirdbest{0.34}
        & 0.58
        & \thirdbest{0.32}
        & \thirdbest{0.33}
        & 0.34
        & \thirdbest{0.49}
        & 20 min
        & RTX 4090 \\

        % & BlobsToSpokes (30k iter)
        % & 0.40
        % & 0.51
        % & 0.27
        % & 0.32
        % & 0.81
        % & 0.46
        % & 0.45
        % & 1.04
        % & 0.65
        % & 0.52
        % & 0.33
        % & 0.59
        % & 0.32
        % & 0.33
        % & 0.34
        % & 0.49
        % & 27 min
        % & RTX 4090 \\

        \midrule

        & CVTSplat
        & \thirdbest{0.35}
        & \best{0.49}
        & \thirdbest{0.30}
        & \best{0.29}
        & \secondbest{0.76}
        & \secondbest{0.45}
        & 0.46
        & \best{0.97}
        & 0.75
        & \best{0.49}
        & 0.38
        & 0.58
        & \best{0.29}
        & \secondbest{0.32}
        & \best{0.31}
        & \secondbest{0.48}
        & \best{2.5 min}
        & RTX 5090 \\

        \bottomrule
    \end{tabular}%
    }

    \vspace{2pt}
    \begin{minipage}{\textwidth}
        \footnotesize
        \textit{Notes:}
        Timings are approximate and are not
        hardware-normalized. GGGS and BlobsToSpokes use the 20k iterations setting.
    \end{minipage}
\end{table*}

\subsection{Training Speed Analysis on DTU}

\begin{table}[!htbp]
\centering
\caption{Training speed analysis on DTU.}
\label{tab:chamfer-ablation-config}
\scriptsize
\setlength{\tabcolsep}{3pt}
\resizebox{\columnwidth}{!}{%
\begin{tabular}{lccccccccccccccccc}
\toprule
Method (\#Sites, \#Optim. steps)
& 24 & 37 & 40 & 55 & 63 & 65 & 69 & 83
& 97 & 105 & 106 & 110 & 114 & 118 & 122
& Mean & Time (s) \\
\midrule

1M Sites, 10K
& \thirdbest{0.38} & 0.54 & \best{0.30} & \secondbest{0.30} & \thirdbest{0.82}
& 0.50 & \secondbest{0.47} & \thirdbest{1.02} & \secondbest{0.75} & 0.53
& \best{0.33} & \thirdbest{0.66} & \secondbest{0.30} & \thirdbest{0.34} & \thirdbest{0.34}
& \secondbest{0.50} & 87 \\

2M Sites, 10K
& \secondbest{0.36} & \thirdbest{0.51} & \best{0.30} & \secondbest{0.30} & \thirdbest{0.82}
& \thirdbest{0.48} & \secondbest{0.47} & \secondbest{1.00} & \best{0.74} & \thirdbest{0.51}
& \best{0.33} & \secondbest{0.65} & \secondbest{0.30} & \secondbest{0.33} & \secondbest{0.33}
& \secondbest{0.50} & 110 \\

4M Sites, 5K
& \secondbest{0.36} & 0.58 & \best{0.30} & \secondbest{0.30} & \best{0.75}
& \secondbest{0.46} & \best{0.46} & \best{0.97} & \best{0.74} & \secondbest{0.50}
& \thirdbest{0.39} & 0.75 & \best{0.29} & \best{0.32} & \best{0.31}
& \secondbest{0.50} & 79 \\

4M Sites, 7.5K
& \best{0.35} & \secondbest{0.50} & \best{0.30} & \secondbest{0.30} & \best{0.75}
& \secondbest{0.46} & \best{0.46} & \best{0.97} & \best{0.74} & \best{0.49}
& \thirdbest{0.39} & \secondbest{0.65} & \best{0.29} & \best{0.32} & \best{0.31}
& \best{0.48} & 115 \\

\midrule

Basic Rasterizer (4M Sites, 10K)
& \best{0.35} & \best{0.49} & \best{0.30} & \best{0.29} & \secondbest{0.76}
& \best{0.45} & \best{0.46} & \best{0.97} & \thirdbest{0.76} & \best{0.49}
& 0.41 & \best{0.58} & \best{0.29} & \best{0.32} & \best{0.31}
& \best{0.48} & 280 \\

Chunked Rasterizer (4M Sites, 10K)
& \best{0.35} & \best{0.49} & \best{0.30} & \best{0.29} & \secondbest{0.76}
& \best{0.45} & \best{0.46} & \best{0.97} & \secondbest{0.75} & \best{0.49}
& \secondbest{0.38} & \best{0.58} & \best{0.29} & \best{0.32} & \best{0.31}
& \best{0.48} & 147 \\

\bottomrule
\end{tabular}%
}
\end{table}

\begin{figure}[!htbp]
    \centering
\includegraphics[width=1.0\linewidth]{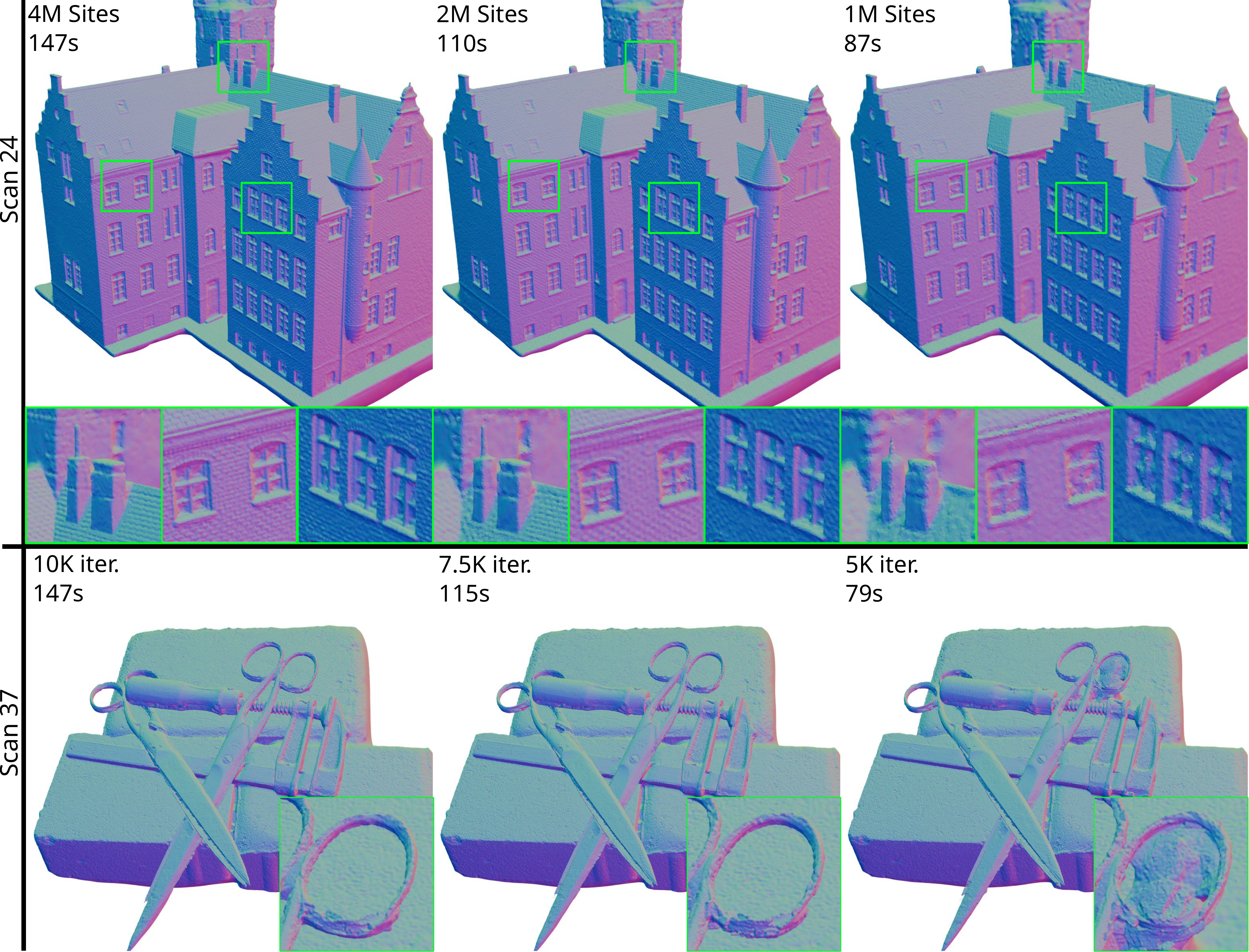}
    \caption{Test with fewer sites or fewer optimization steps. Top: Using fewer sites produces a predictable loss of geometric details with faster optimization. Bottom: Our model converges after 10K training steps, with marginal improvement beyond 7.5K steps.}
    \label{fig:ablations}
\end{figure}

We evaluate in table \ref{tab:chamfer-ablation-config} the two main configuration hyperparameters of our approach: the number of sites and the number of training iterations. Reducing either of these produces a small decrease in accuracy but lowers training time substantially. Qualitative results are shown in figure \ref{fig:ablations}. 

Next, using the chunked rasterizer described in section \ref{subsec:chunked_rasterizer} has the largest impact on runtime: the optimization takes $147$s with it, compared to $280$s using a baseline that assigns a single GPU thread per pixel similarly to 3DGS \cite{kerbl20233d}. This almost gives a $2\times$ speedup on the end-to-end optimization. Both rasterizers achieve nearly the same Chamfer distance and the remaining discrepancy can likely be attributed to differences in floating-point arithmetic.

\subsection{Ablations on DTU}

\begin{figure}[!htbp]
    \centering
\includegraphics[width=1.0\linewidth]{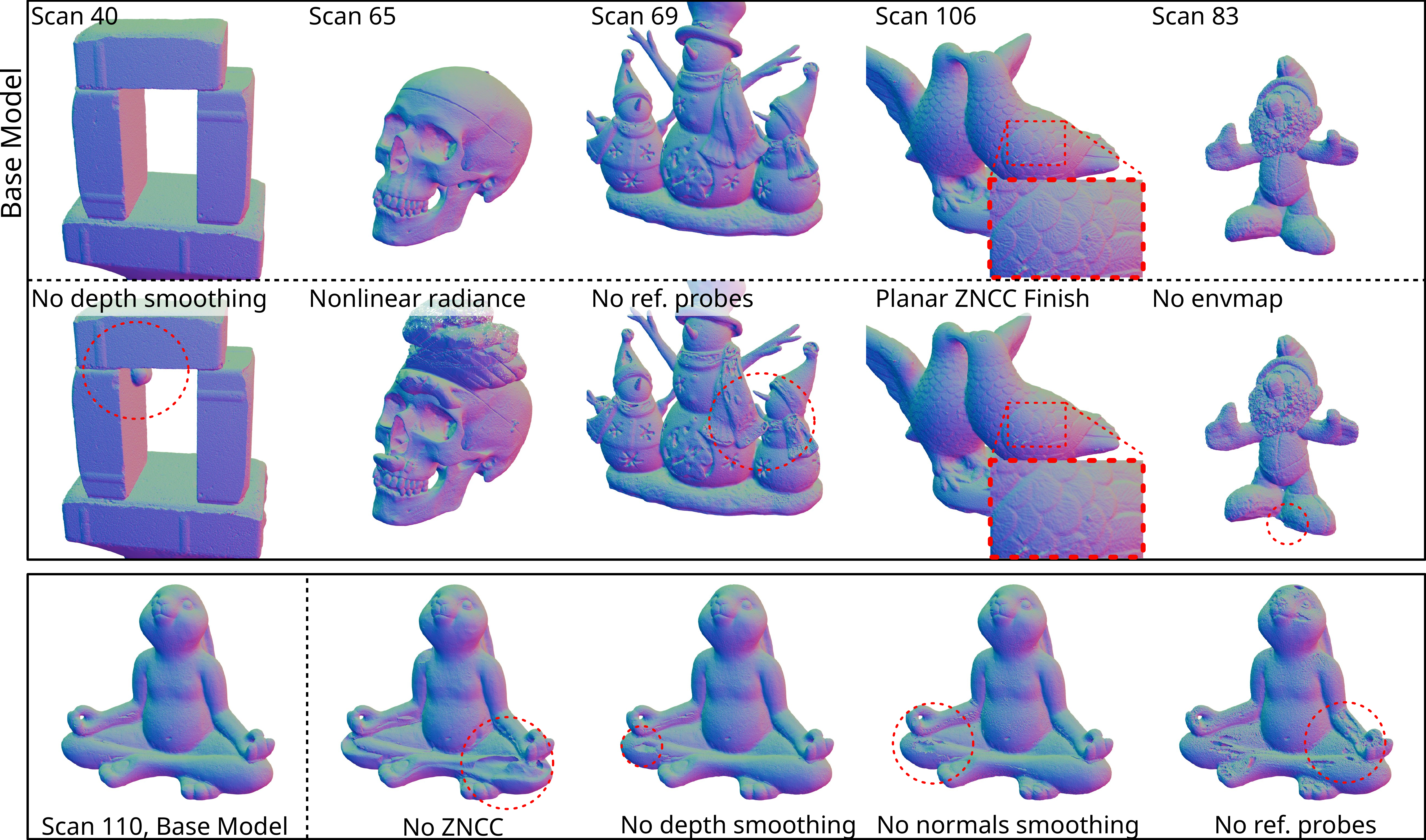}
    \caption{We remove one loss on each example to ablate our model. Bottom: The multi-view (ZNCC), depth and normal losses as well as the reflection probes are important for geometric accuracy. Top: Training with a linear radiance avoids a failure case on scan 65. Not replacing the planar ZNCC with a nonplanar variant in the last phase of training suppresses fine details (scan 106). The envmap alleviates silhouette artifacts from inaccurate masks (scan~83).}
    \label{fig:ablations2}
\end{figure}

We ablate our losses and regularization terms in table \ref{tab:chamfer-ablation-losses}, ordered by decreasing importance. Figure \ref{fig:ablations2} compares against the base model qualitatively. Training using a linear radiance tone-mapped after rasterization, compared to a typical sigmoid-bounded nonlinear radiance, avoids a failure case on scan 65 due to large patches of saturated pixels, but makes little difference otherwise. Removing the multi-view loss (No ZNCC) or the depth smoothing loss has a strong impact overall. Setting $a=0.05$ in the biased CVT loss (eq. \ref{eq:CVTLoss}), instead of $a=0.10$ prevents the cells from shrinking sufficiently which reduces accuracy but does not create hard failure cases. The reflection probes are important for recovering smooth geometry in specular regions (scan 110, 69). Adding an environment map helps compensating for slight inaccuracies in the supervision masks (scan 83). The rest of the components have lower importance: the SSIM, color correction curves and view probes are added for robustness against inconsistent illumination conditions. Removing one of them does not affect geometry meaningfully. Finishing training with the non planar ZNCC term instead of the planar ZNCC gives a qualitative improvement in fine details (scan 106). Finally, removing the normal smoothing term (eq. \ref{eq:normal_smoothing}) slightly improves the Chamfer distance compared to our base model but also introduces artifacts (scan 110). The gradient modulation (eq. \ref{eq:grads_mod}) does not change the end result but makes training more stable.

\begin{table}[!htbp]
\centering
\caption{Ablation study on DTU.}
\label{tab:chamfer-ablation-losses}
\scriptsize
\setlength{\tabcolsep}{3pt}
\resizebox{\columnwidth}{!}{%
\begin{tabular}{lcccccccccccccccc}
\toprule
Method
& 24 & 37 & 40 & 55 & 63 & 65 & 69 & 83
& 97 & 105 & 106 & 110 & 114 & 118 & 122
& Mean \\
\midrule

No Linear Radiance
& \secondbest{0.35} & \best{0.49} & \secondbest{0.30} & \best{0.29} & 0.79
& 1.78 & \best{0.44} & 0.99 & 0.84 & 0.50
& \best{0.32} & 0.59 & \secondbest{0.29} & \secondbest{0.32} & \secondbest{0.31}
& 0.57 \\

No ZNCC
& 0.40 & 0.57 & \thirdbest{0.33} & \thirdbest{0.31} & 0.80
& 0.52 & 0.52 & 1.05 & 0.83 & 0.54
& \thirdbest{0.38} & 1.07 & \secondbest{0.29} & 0.35 & \thirdbest{0.32}
& 0.55 \\

No Depth Smoothing
& 0.44 & \secondbest{0.50} & 0.46 & \best{0.29} & \secondbest{0.75}
& \secondbest{0.45} & \secondbest{0.46} & \thirdbest{0.97} & \thirdbest{0.75} & 0.50
& 0.50 & 0.70 & \secondbest{0.29} & 0.36 & \secondbest{0.31}
& 0.52 \\

Biased CVT ($a=0.05$)
& \thirdbest{0.36} & \thirdbest{0.51} & 0.34 & \secondbest{0.30} & 0.81
& 0.50 & 0.48 & 1.02 & 0.77 & 0.52
& \secondbest{0.34} & 0.80 & \thirdbest{0.30} & 0.34 & 0.34
& 0.52 \\

No Ref. Probes
& \secondbest{0.35} & 0.56 & \secondbest{0.30} & \best{0.29} & \thirdbest{0.76}
& \secondbest{0.45} & 0.50 & \thirdbest{0.97} & 0.82 & \thirdbest{0.49}
& \thirdbest{0.38} & 0.81 & \thirdbest{0.30} & \secondbest{0.32} & \secondbest{0.31}
& 0.51 \\

No Envmap
& \secondbest{0.35} & 0.52 & \best{0.29} & \best{0.29} & 0.80
& \best{0.44} & \best{0.44} & 1.12 & 0.76 & 0.53
& 0.39 & 0.64 & \best{0.28} & \best{0.31} & \best{0.30}
& \thirdbest{0.50} \\

\midrule

Planar ZNCC Only
& \secondbest{0.35} & \thirdbest{0.51} & \secondbest{0.30} & \best{0.29} & \secondbest{0.75}
& \secondbest{0.45} & \secondbest{0.46} & \secondbest{0.96} & \thirdbest{0.75} & 0.50
& 0.39 & 0.60 & \thirdbest{0.30} & \secondbest{0.32} & \secondbest{0.31}
& \secondbest{0.48} \\

No Color Correction
& \secondbest{0.35} & \best{0.49} & \secondbest{0.30} & \best{0.29} & \secondbest{0.75}
& \best{0.44} & \secondbest{0.46} & \thirdbest{0.97} & 0.76 & 0.50
& 0.41 & 0.59 & \secondbest{0.29} & \secondbest{0.32} & \secondbest{0.31}
& \secondbest{0.48} \\

No View Probes
& \secondbest{0.35} & \best{0.49} & \secondbest{0.30} & \best{0.29} & \secondbest{0.75}
& 0.47 & \thirdbest{0.47} & \thirdbest{0.97} & \best{0.68} & 0.50
& 0.40 & \thirdbest{0.58} & \secondbest{0.29} & \secondbest{0.32} & \secondbest{0.31}
& \secondbest{0.48} \\

No SSIM
& \secondbest{0.35} & \thirdbest{0.51} & \secondbest{0.30} & \best{0.29} & \secondbest{0.75}
& \thirdbest{0.46} & \secondbest{0.46} & \thirdbest{0.97} & \secondbest{0.70} & 0.51
& \best{0.32} & 0.60 & \thirdbest{0.30} & \thirdbest{0.33} & \secondbest{0.31}
& \secondbest{0.48} \\

No Gradient Mod.
& \best{0.34} & \best{0.49} & \best{0.29} & \best{0.29} & \secondbest{0.75}
& \best{0.44} & \secondbest{0.46} & 0.99 & 0.79 & \secondbest{0.48}
& 0.41 & \best{0.52} & \best{0.28} & \best{0.31} & \best{0.30}
& \secondbest{0.48} \\

No Normal Smoothing
& \secondbest{0.35} & \secondbest{0.50} & \best{0.29} & \best{0.29} & \best{0.74}
& \secondbest{0.45} & 0.49 & \best{0.93} & \thirdbest{0.75} & \best{0.46}
& 0.43 & \secondbest{0.54} & \secondbest{0.29} & \best{0.31} & \best{0.30}
& \best{0.47} \\

\midrule
Base Model
& \secondbest{0.35} & \best{0.49} & \secondbest{0.30} & \best{0.29} & \thirdbest{0.76}
& \secondbest{0.45} & \secondbest{0.46} & \thirdbest{0.97} & \thirdbest{0.75} & \thirdbest{0.49}
& \thirdbest{0.38} & \thirdbest{0.58} & \secondbest{0.29} & \secondbest{0.32} & \secondbest{0.31}
& \secondbest{0.48} \\

\bottomrule
\end{tabular}%
}
\end{table}

\subsection{KNN Efficiency}

We measure the speed of our KNN (K=32) implementation against five baselines on the 4M points DTU checkpoints and present the results in table \ref{tab:knn_benchmark}. For our approach, we distinguish between the case where the ball query search radius is already known (w. warmup) compared to when it is unknown which takes approximately $3\times$ longer. In practice, most of the sites can rely on the search radius used in the KNN lookup of the previous training iteration. 

\begin{table}[!htbp]
    \centering
    \caption{KNN timings (ms) on our DTU point clouds with 4M points. The * symbol denotes CPU operations.}
    \label{tab:knn_benchmark}
    \resizebox{\columnwidth}{!}{%
    \begin{tabular}{lcccccc}
\toprule
        & \makecell{CVTSplat\\ with/without warmup} & \makecell{RestrictedVoronoi\\ \cite{basselin2021restricted}} & \makecell{CuPy-KNN\\ \cite{cupy_knn} } & \makecell{Torch-kdtree\\ \cite{torch_kdtree} } & \makecell{Sklearn\\ \cite{scikit-learn} } & \makecell{TrueKNN \\ \cite{nagarajan2023rt} } \\
\midrule
        Build Tree & \textbf{1.3} & <1 & 17.3 & 6742* & 1925* & - \\
        Query Tree & \textbf{13.7} / 46.8 & 97.9 & 720 & 1026 & 25630* & - \\
        Total & \textbf{15.0} / 48.1 & 97.9 & 737 & 7768 & 27556 & 28543 \\
\bottomrule
    \end{tabular}%
    }
\end{table}

MeshlessVoronoi \cite{ray2018meshless} and RestrictedVoronoi \cite{basselin2021restricted} use the same strategy so we only discuss \cite{basselin2021restricted} which was slightly faster. The input points are partitioned on a dense voxel grid as a pre-process. Then, each query point visits the voxels in concentric rings and keeps track of the KNN using a heap in shared memory. Importantly, we doubled their hard-coded maximum search radius to accommodate our adaptive point distributions, otherwise these baselines fail to gather all requested neighbors.

CuPy-KNN \cite{cupy_knn} is a re-implementation of \cite{jakob2021optimizing}, based on a BVH and using a heap stored in the registers. Torch-kdtree \cite{torch_kdtree} works similarly to Cupy-KNN but builds its acceleration structure on the CPU. Overall, the heap-based approaches are much less efficient than our gather-then-sort approach, even more so when temporal consistency is leveraged. 
TrueKNN \cite{nagarajan2023rt} relies on the RT cores of the GPU to benefit from hardware-accelerated BVH traversal. This strategy is effective for uniform point distributions but under performs on our point clouds since several retries are needed for low-density regions. We also include scikit-learn \cite{scikit-learn} as a CPU reference.

\subsection{Voronoi and Delaunay Efficiency}

\begin{table}[!htbp]
\centering
\caption{
    Voronoi diagram construction timings (ms) on our trained DTU point clouds (4M points).
}
\label{tab:dtu_voronoi_benchmark}
\setlength{\tabcolsep}{3.5pt}
\renewcommand{\arraystretch}{1.05}

\resizebox{\linewidth}{!}{%
\begin{tabular}{lcccccc}
\toprule
    & \makecell{CVTSplat\\(Ours)}    & \makecell{MeshlessVoronoi\\\cite{ray2018meshless}}   & \makecell{RestrictedVoronoi\\\cite{basselin2021restricted}}   & \makecell{Paragram\\\cite{taveira2026scalable}}   & \makecell{gDel3D\\\cite{Cao2014gDel3D}}  & \makecell{RadFoam\\\cite{govindarajan2025radiant}}  \\
\midrule
 KNN & \textbf{15.0} & 126.6 & 97.9 & -  & - & - \\
 %\midrule
 Adjacency   & \textbf{11.9} & \multirow{2}{*}{44.5}  & \multirow{2}{*}{40.5}   & 275.5      & 617.4    & 18995   \\
 Centroids   & \textbf{3.0}  &  &  & -          & -        & -         \\
 \midrule
 Total       & \textbf{29.9}  & 171.0      & 138.4 & 275.5  & 617.4    & 18995   \\
\bottomrule
\end{tabular}
}
\end{table}
\begin{figure}[h]
    \centering
    \includegraphics[width=1.0\linewidth]{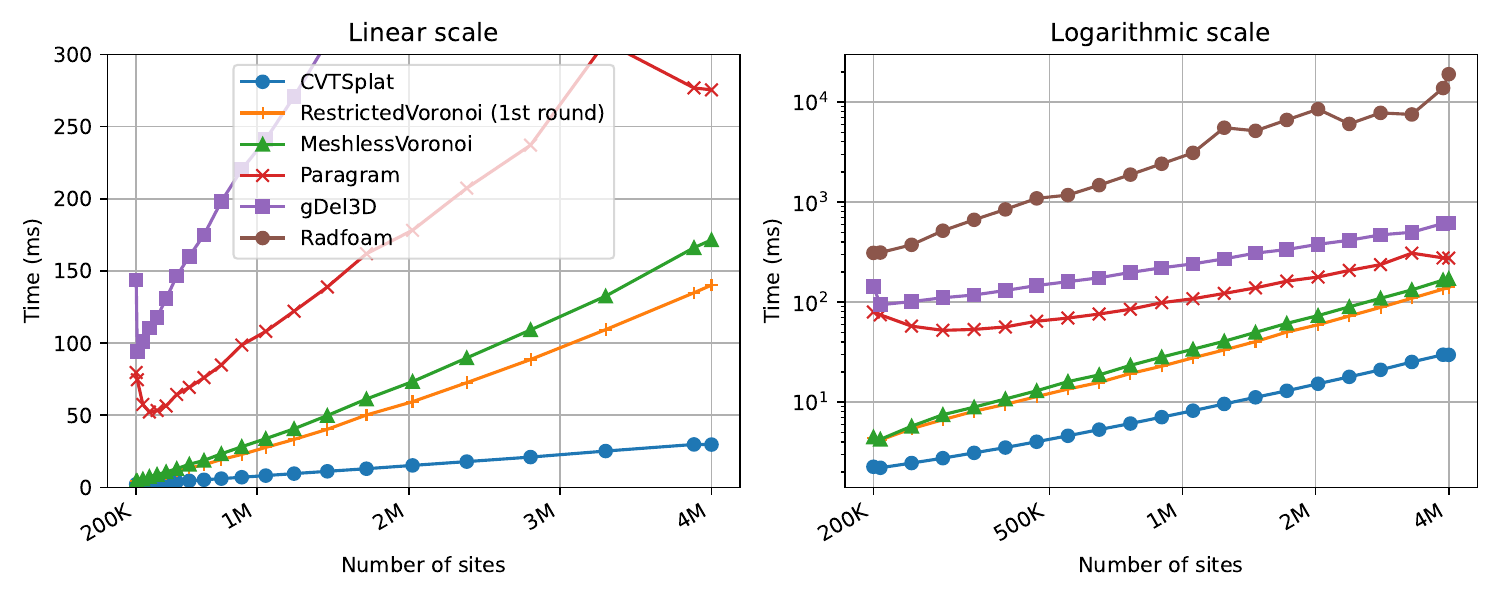}
    \caption{Timings for Voronoi diagram construction, evaluated on DTU checkpoints saved every 500 optimization steps. Reported numbers include KNN and centroids computations if applicable.}
    \label{fig:timings}
\end{figure}

We compare the speed of our algorithm for Voronoi diagram construction against five GPU baselines in table \ref{tab:dtu_voronoi_benchmark} and in figure \ref{fig:timings}, on the 4M points DTU checkpoints. Paragram \cite{taveira2026scalable}, gDel3D \cite{Cao2014gDel3D} and RadiantFoam \cite{govindarajan2025radiant} only output the Voronoi adjacency graph. Closer to our approach, MeshlessVoronoi \cite{ray2018meshless} and RestrictedVoronoi \cite{basselin2021restricted} produce a list of KNN and also compute cell centroids. We ran both of them with single-precision arithmetic and K=32 neighbors, equivalently to our setting. In the case of RestrictedVoronoi \cite{basselin2021restricted}, we only report the first round of KNN gathering which corresponds to what we are doing. Subsequent retries with larger K for incomplete cells increase their runtime by about $1.5\times$. The reported timing for our approach assumes the KNN search radius is known, to match the typical case encountered during optimization. Overall, our approach is consistently $5\times$ faster than \cite{ray2018meshless} and $10\times$ faster than \cite{taveira2026scalable} in this context, and scales linearly.

\begin{table}[h]
\centering
\caption{Mesh extraction timings on DTU.}
\label{tab:mesh_extraction}
\resizebox{\columnwidth}{!}{%
\begin{tabular}{lccc}
\toprule
      & \makecell{Adjacency $\rightarrow$ Del.}
      & \makecell{Marching Tet.}
      & Total \\
\midrule
All cells & 11.5 ms & 4.0 ms & 15.5 ms \\
\makecell[l]{Zero-crossing only}
    & \textbf{4.5 ms}
    & \textbf{0.6 ms}
    & \textbf{5.1 ms} \\
\bottomrule
\end{tabular}%
}
\end{table}
For mesh extraction, we convert the adjacency graph into a Delaunay triangulation (sec. \ref{sec:delaunay_triangulation}) and subsequently apply marching tetrahedra on the smoothed SDF $\hat{f}$. It is sufficient to only generate the tetrahedra that cross the zero level set, which divides the total mesh extraction time by $3\times$ according to table \ref{tab:mesh_extraction}. The zero level set is extracted as a triangle soup for real-time visualization during training. The duplicate vertices are merged on the CPU before outputting the final mesh on disk, which takes two seconds in total.

\section{Limitations and Future Work}

A limitation of our Voronoi diagram construction algorithm is its approximate nature, which makes it unsuitable for applications requiring exact adjacencies. Iterative retries with a larger number of neighbors in the spirit of Basselin et al.~\cite{basselin2021restricted} could recover the missing adjacencies. A fallback to double precision could also improve robustness in degenerate configurations. 

As evaluated in the supplementary material, we find the proportion of incomplete cells to be low in practice for our choice of hyperparameters. More aggressive adaptivity could be balanced using a larger number of neighbors but at an increased cost. Deriving theoretical upper bounds on the required number of neighbors based on cell regularity~\cite{Hudson2006SparseVR} could be useful but may be inapplicable in our case since we make no guarantee on sites spacing. 

Extending our adaptive discretization to open scenes is an interesting direction for future research. The main obstacles are our naive sphere initialization and the potential lack of scalability for representing geometric and photometric details over large spatial extents. The former could be addressed using monocular priors~\cite{li2026geosvr} and the latter could benefit from adaptive densification based on image content \cite{kerbl20233d, govindarajan2025radiant}, or primitives with greater expressivity \cite{taveira2026vorotracing}.

\section{Conclusion}

We have presented CVTSplat, an approach for multi-view reconstruction combining adaptive Voronoi tessellations with differentiable implicit surface modeling. Our centroidal regularization plays a central role: beyond producing well-shaped cells, it enables us to specialize several components of the pipeline for efficiency. In particular, we introduced a Voronoi diagram construction algorithm that exploits the regularity of the tessellation. We further tailored neighbor queries, centroid computation, rasterization and mesh extraction to our representation. Together, these developments yield a highly efficient pipeline that achieves geometric accuracy competitive with the state-of-the-art on two object-level reconstruction benchmarks.

\section*{Acknowledgments}

This work was partly supported by AMEDJP26wm0625404 and JSPS/KAKENHI JP26K14883.

%% file: main_suppmat.tex
\pagebreak

\section{Overview}

This supplementary document provides additional implementation details, experiments, derivations and proofs. We report the optimization schedule (section \ref{sec:hyperparameters}) followed by statistics on our adaptive Voronoi tessellation: study of mesh defects (section \ref{sec:mesh_defects}), accuracy of cell moments computation (section \ref{sec:moments_accuracy}), frequency of incomplete cells and relationship between cell size and SDF (section \ref{sec:cell_statistics}). Evaluations of rendering speed, photometric accuracy and model size are available in sections \ref{sec:rendering_speed}, \ref{sec:photo_eval} and \ref{sec:model_size} respectively. Section \ref{sec:mvloss} gives details regarding the multi-view loss and section \ref{sec:derivations_and_proofs} concludes with proofs omitted from the main paper.

\section{Hyperparameters Scheduling} \label{sec:hyperparameters}

Our optimization schedule is detailed in table \ref{tab:default-schedule}.

\begin{table}[!htbp]
\centering
\small
\caption{Optimization schedule. Array values are linearly interpolated throughout training.}
\label{tab:default-schedule}
\begin{tabular}{@{}p{0.96\linewidth}@{}}
\hline

\textbf{Training phases.}
10,000 iterations: 400 initialization, 9200 main training, and 400 cleanup. No sites are added during initialization or cleanup. \\

\textbf{Point population.}
200k initial sites up to 4M sites. Densification every 10 iterations. \\

\textbf{Data and optimization.}
4 image levels of detail, batch size: 4 images per batch using gradient accumulation, learning rate
\texttt{[0.025, 0.001]}. \\

\textbf{Structure updates.}
KNN every 10 iterations, Voronoi diagram construction frequency \texttt{[1, 2, 5]}. \\

\textbf{Geometric regularization.}
$\mathcal{L}_{\text{SDF}}=$\texttt{[0.4, 0.1]}, $\mathcal{L}_{\mathbf{n}}=$\texttt{[0.05, 0.1]}, $\mathcal{L}_{\text{Eik}}=$\texttt{[0.005]}, depth smoothing $\mathcal{L}_{\text{D}}=$\texttt{[0.5]}. \\

\textbf{CVT Densification.} 
$\mathcal{L}_{\text{CVT}}=1$, $a=$\texttt{[0.05, 0.10]}, $b=$\texttt{[0.01, 0.001, 0.0001]}. \\

\textbf{Image losses.}
$\mathcal{L}_{\text{RGB}}=$\texttt{[5, 10]}, $\mathcal{L}_{\text{DSSIM}}=$\texttt{[1, 2]}, $\mathcal{L}_{\text{MV}}=$\texttt{[0.1, 10]}, $\mathcal{L}_{\text{Mask}}=$\texttt{[5, 0.1]}. \\

\textbf{Appearance.} 
NeuS sharpness $s=$\texttt{[5, 15, 30, 60, 120]}, features smoothing $\mathcal{L}_{\mathbf{F}^s}=$\texttt{[0.05]}, probes smoothing $\mathcal{L}_{\text{probes}}=$\texttt{[0.1, 1]}. \\
$1024^2$ background envmap, two $32^3$ spherical-harmonic probe grids.\\
\hline
\end{tabular}
\end{table}

\section{Mesh Defects Analysis} \label{sec:mesh_defects}

Table \ref{tab:mesh_defects} reports the average number of topological errors in our extracted meshes. We count the number of non-manifold edges, \ie{} edges having more than two incident triangles, and boundary edges that have a single incident triangle. The implicit surface touches the bounding box walls in some scenes, which creates legitimate boundary edges. Therefore, we only count the ones within $95\%$ of the bounding box that actually correspond to small holes in the extracted meshes. Missing Voronoi adjacencies can prevent the formation of tetrahedra and produce small holes, where a single triangle is missing in general. Non-manifold edges may result from inconsistent tetrahedron selection near degenerate configurations or from incomplete Voronoi neighbor lists. Overall, these defects are infrequent relative to the number of mesh triangles and could potentially be addressed by topological post-processing.

\begin{table}[!htbp]
\centering
\caption{Average mesh extraction defects by dataset.}
\label{tab:mesh_defects}
\resizebox{\columnwidth}{!}{%
\begin{tabular}{lccc}
\hline
Dataset & Non-manifold edges & Boundary edges & Triangles\\
\hline
DTU        & 2.7 & 2.3 & 6.4M\\
BlendedMVS & 2.4 & 1.4 & 6.7M\\
\hline
\end{tabular}%
}
\end{table}

\section{Cell Moments Estimation} \label{sec:moments_accuracy}

\begin{figure}[!htbp]
    \centering
    \includegraphics[width=1.0\linewidth]{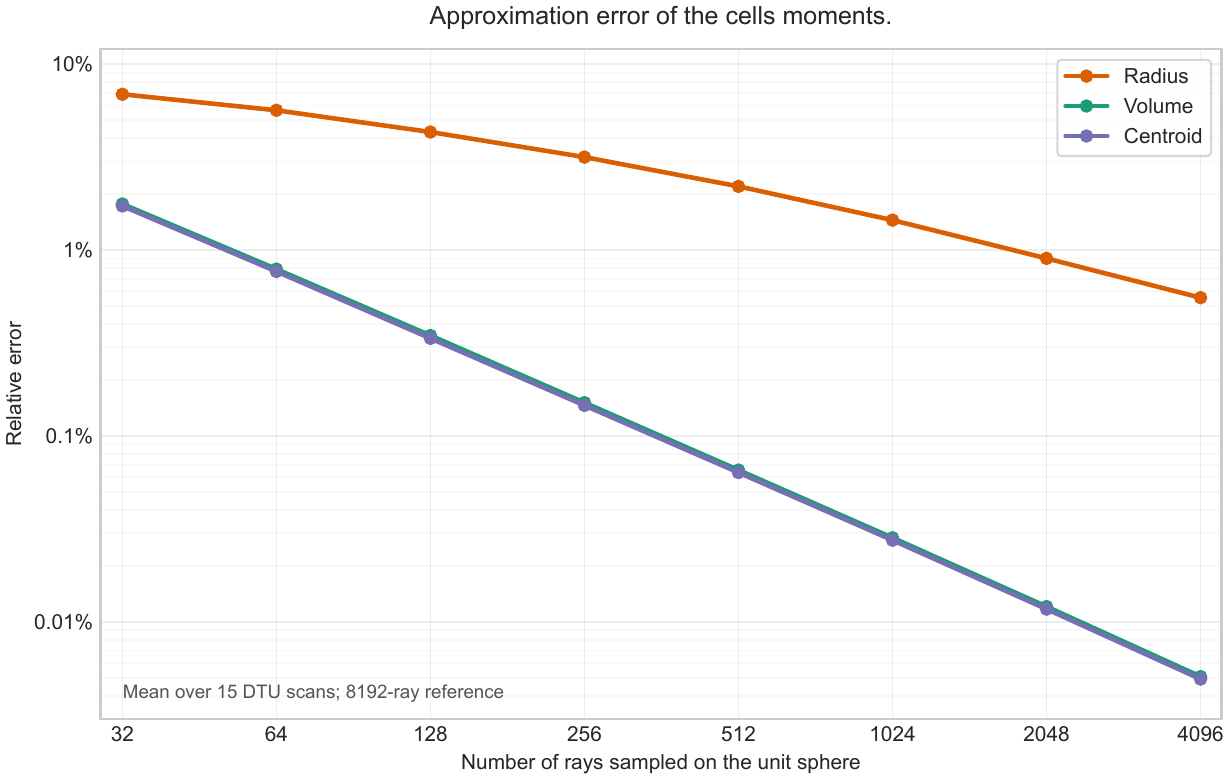}
    \caption{Relative error curves when computing cell volume, centroid and bounding sphere radius using a varying number of ray samples. The volume and centroid error curves are nearly overlapped in the plot.}
    \label{fig:moments_accuracy}
\end{figure}

We evaluate in Figure \ref{fig:moments_accuracy} the accuracy of the quadrature formulas for cell volume and centroid using varying numbers of rays sampled on a Fibonacci sphere lattice. Errors are relative and computed with respect to a reference using $8192$ rays. Volume and radius errors are relative to their reference values while the centroid error corresponds to the centroid displacement
divided by the reference cell radius. The radius is estimated from the maximum sampled distance to the cell boundary over the set of ray directions. The error curves for the volume and centroid are nearly overlapped and decrease as $1/N$, where $N$ is the number of rays. The radius error decreases more slowly, approximately as $1/\sqrt{N}$.

With our default of 64 rays, the mean errors are $0.79\%$ for the volume, $0.77\%$ for the centroid, and $5.65\%$ for the radius. The cell radius mainly serves to define an approximate bounding sphere during rasterization. A more accurate value could be computed using the maximum distance to the circumcenters of incident tetrahedra (\ie{} the Voronoi vertices), but that would require computing the Delaunay triangulation from the adjacency graph every time it is updated which would add some overhead. Our approximate bounding spheres only clip the furthest corners of the cells, that would not contribute significantly to the volume integral if included.

\section{Distribution of cells} \label{sec:cell_statistics}

\begin{figure*}[!htbp]
    \centering
    \includegraphics[width=1.0\linewidth]{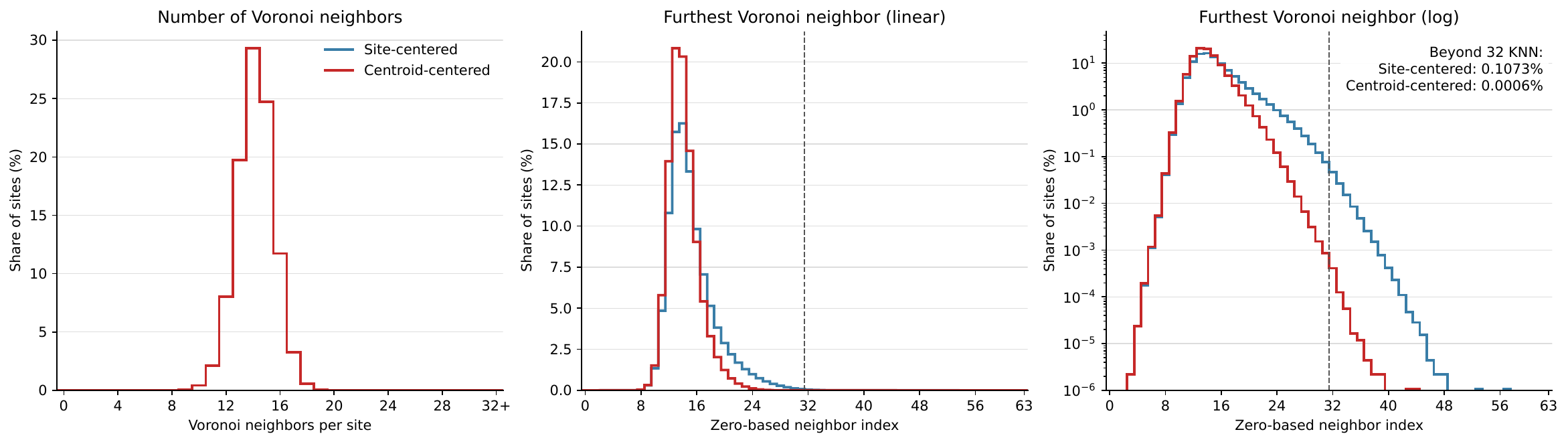}
    \caption{Histograms pooled over all 23 DTU and BlendedMVS scenes (92 million sites). The blue one queries KNN at each site while the red one queries at the cell centroid computed at the previous update. \textbf{Left}: Number of Voronoi neighbors per site. The mean is $\approx14.2$, corresponding to $14$ bisector planes contributing to each cell on average. \textbf{Middle and Right}: zero-based index of the last detected Voronoi neighbor in the KNN list, with linear and logarithmic frequency scales respectively. For $\approx60\%$ of sites, all Voronoi neighbors lie among the first 16 nearest neighbors. The frequency decreases roughly exponentially beyond index 15, with $\approx0.1\%$ of cells not finding their last Voronoi neighbor within the 32-KNN list with the base strategy. Using shifted ball queries reduces the frequency of incomplete cells by a factor of 170 in comparison ($\approx600$ cells in total out of 92M). }
    \label{fig:neighbor_histograms}
\end{figure*}

\begin{figure*}[!htbp]
\centering
\includegraphics[width=\linewidth]{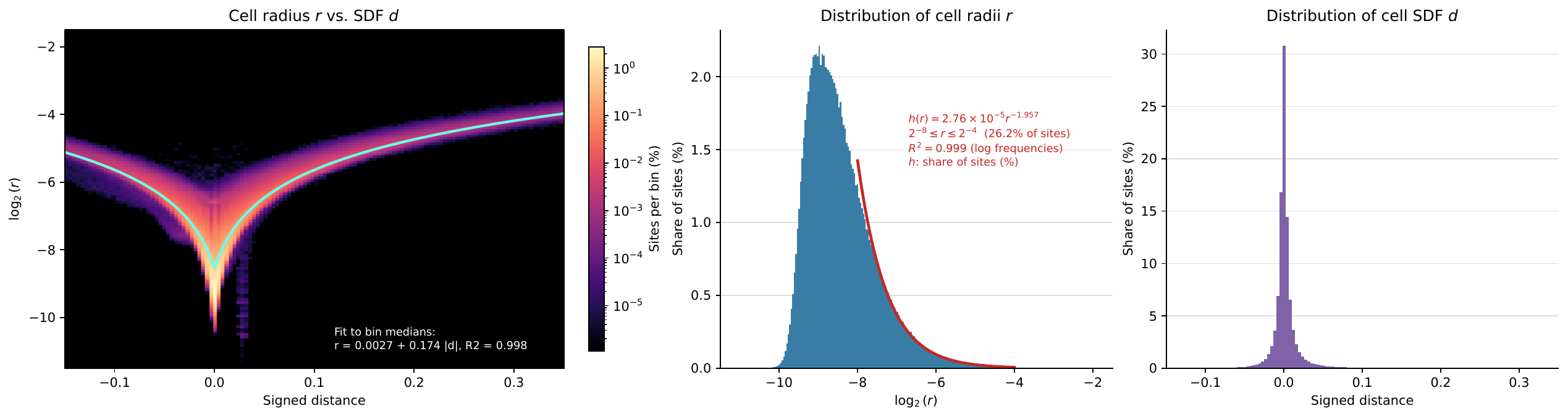}
\caption{Cell-radius statistics averaged over all scenes. \textbf{Left}: the distribution of cell radius $r$ against SDF $d$. The cyan curve fits the median radius in each distance bin, showing approximately linear growth with $|d|$. Its similar branches on either side of zero indicate comparable cell size inside and outside the surface at equal distance. \textbf{Middle}: the distribution of $\log_2(r)$. The red curve fits a power-law tail over $2^{-8}\leq r\leq2^{-4}$, containing 26.2\% of sites. Within this range, each doubling of $r$ corresponds to approximately $3.9\times$ fewer sites per histogram bin ($h(r)/h(2r) \approx 3.9$). \textbf{Right}: The number of cells decreases very quickly away from the implicit surface. }
\label{fig:cell_radius_distribution}
\end{figure*}

Figure \ref{fig:neighbor_histograms} presents statistics about the number of Voronoi neighbors and their distribution within the KNN list. This test used 64 nearest neighbors instead of 32 for validation. A small proportion of cells  has Voronoi neighbors beyond the 32nd nearest neighbor ($\approx0.1\%$). Extracting the KNN at the centroid decreases that rate substantially ($\approx0.0006\%$) but has little importance on multi-view optimization in practice. Incomplete cells are typically located away from the implicit surface and therefore do not participate in volume rendering or mesh extraction. Nevertheless, we use the offset KNN strategy by default as it provides lower error rates at no additional cost.

Figure \ref{fig:cell_radius_distribution} explores the relationship between cell size and signed distance induced by our adaptive CVT. The bounding box used for training spans the unit cube $[-1, +1]^3$, the SDF and cell radius are given in this reference frame. We find an approximately affine relationship between cell radius and SDF, with most of the cells located close to the implicit surface.

\begin{figure*}[!htbp]
    \centering
    \includegraphics[width=1.0\linewidth]{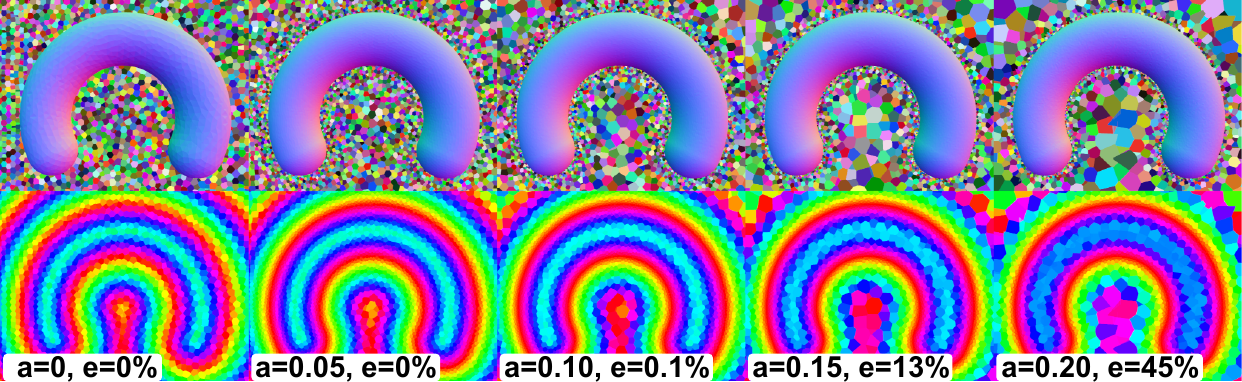}
    \caption{Adaptive discretization of a known SDF. \textbf{Top}: 2D slice with random colors and superimposed extracted mesh. \textbf{Bottom}: 2D slice with SDF mapped to hue. We fit the SDF using $N=10^5$ sites and vary the parameter $a$ in the CVT loss. The mesh approximates the zero level set more accurately for $a > 0$. The proportion of cells $e$ with  missing bisector planes increases sharply for $a>0.1$ when using 32 Euclidean nearest neighbors. Missing bisector planes can cause cells to clearly overlap for larger values of $a$ (see $a=0.2$, center of the horseshoe in the top panel).}
    \label{fig:adaptive_discretization}
\end{figure*}

Figure \ref{fig:adaptive_discretization} illustrates the sensitivity of the point distribution to the hyperparameter $a$ in the CVT loss.

\section{Rendering speed} \label{sec:rendering_speed}

\begin{table}[!htbp]
\centering
\caption{Average projected cell radius and rendering speed
(mean $\pm$ standard deviation across scenes).}
\label{tab:cvtsplat-radius-render-time}
\resizebox{\columnwidth}{!}{%
\begin{tabular}{lcccc}
\toprule
Dataset & Image size & Cell radius (px)
      & Render time (ms) & FPS \\
\midrule
DTU        & $1600 \times 1200$ & $1.55 \pm 0.31$
         & $3.63 \pm 0.29$ & $277 \pm 21$ \\
BlendedMVS & $768 \times 576$ & $1.21 \pm 0.15$
         & $2.40 \pm 0.21$ & $419 \pm 36$ \\
\bottomrule
\end{tabular}%
}
\end{table}

Table \ref{tab:cvtsplat-radius-render-time} reports the average size of the primitives once projected to the training views. Each primitive is weighted by its volume rendering weight to compute this metric. We also report the time taken to render an image from the dataset, averaged over all viewpoints and scenes. The timings are measured on an RTX 5090 GPU. The renderings are written to an offscreen buffer so a small additional overhead would be needed in practice to display the result.

\section{Photometric Evaluation} \label{sec:photo_eval}

We evaluate the photometric accuracy of our approach in table \ref{tab:image-metrics}, on all available viewpoints and within the silhouette masks eroded by two pixels. All tested methods were trained using all viewpoints so this experiment measures fitting error rather than novel view synthesis fidelity. PGSR~\cite{chen2024pgsr} and GGGS~\cite{Zhang2026GeometryGrounded} train on cropped images so we evaluate our method in the same setting for consistency. PGSR and GGGS used half-resolution images on DTU, following their training setup, so we upscale their renderings before evaluation. Our method achieves an image quality comparable to PGSR but slightly below GGGS, with a larger gap on BlendedMVS.

\begin{figure}[!htbp]
    \centering
    \includegraphics[width=1.0\linewidth]{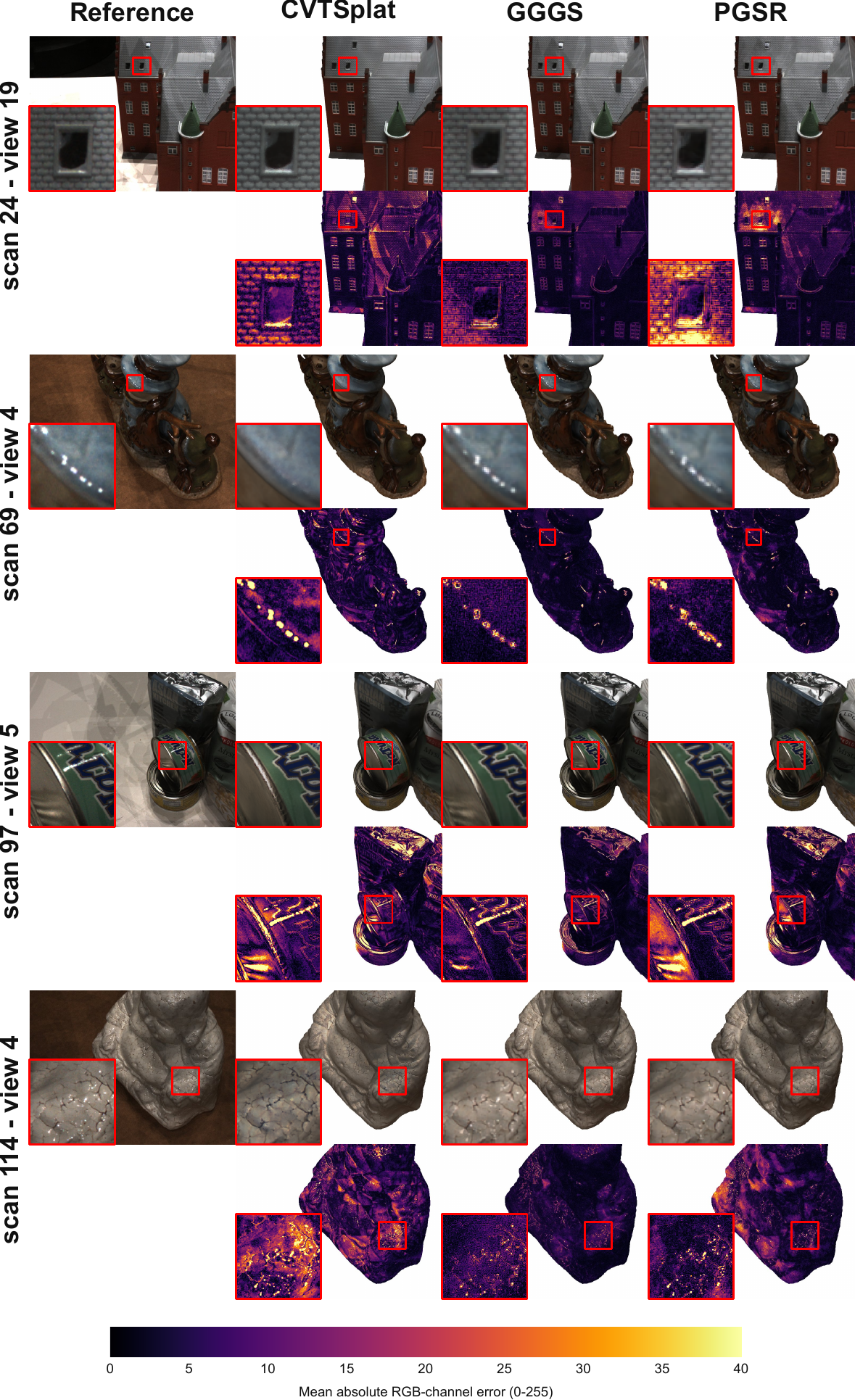}
    \caption{Qualitative comparison of renderings on DTU. Our approach can represent high-frequency texture patterns (scan 24, house roof) but misses some specularities compared to GGGS and PGSR. }
    \label{fig:dtu_qualitative_renders_compressed}
\end{figure}

\begin{figure}[!htbp]
    \centering
    \includegraphics[width=1.0\linewidth]{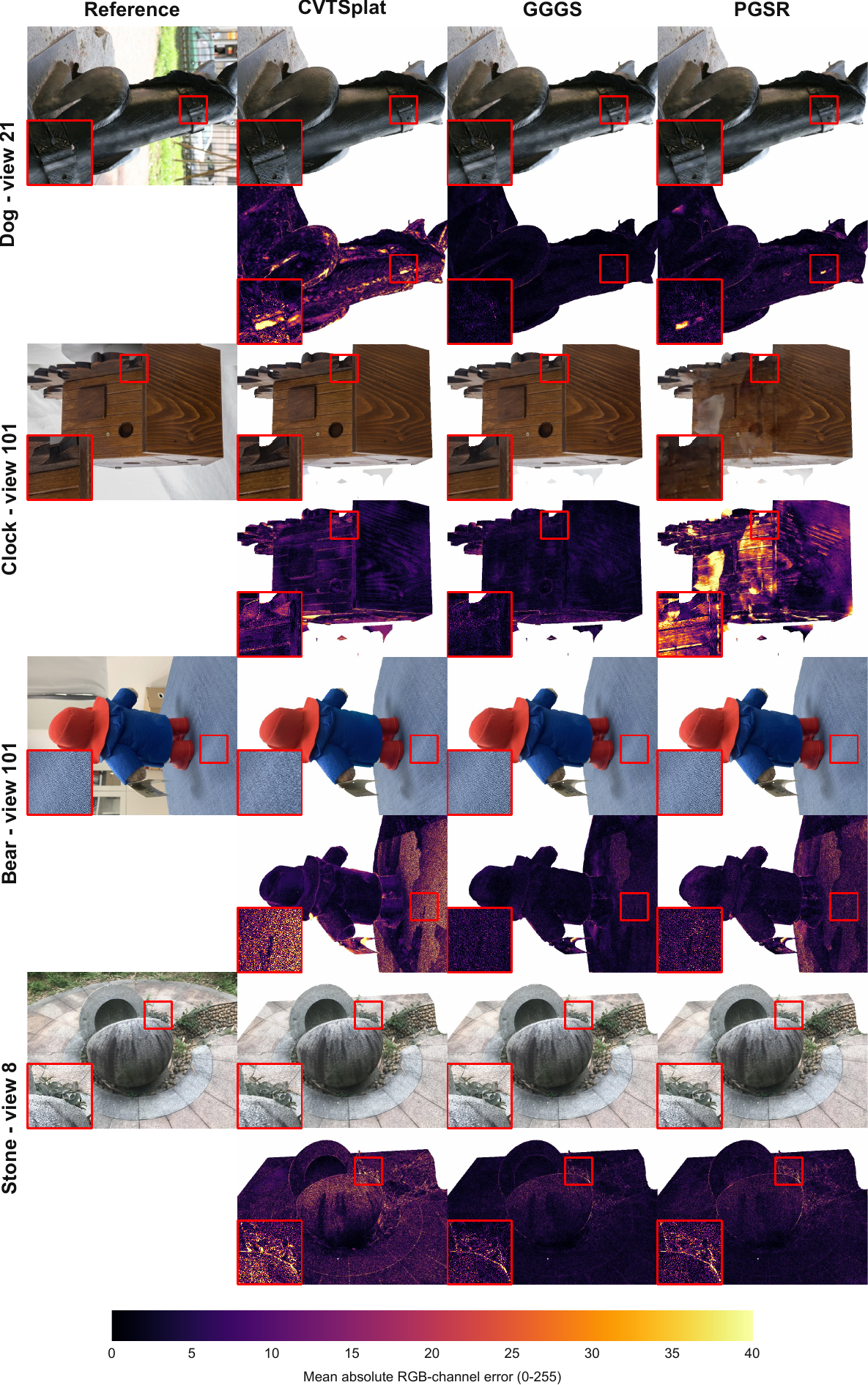}
    \caption{Qualitative comparison of renderings on BlendedMVS. PGSR failed to converge on the clock scene. }
    \label{fig:bmvs_qualitative_renders_compressed}
\end{figure}

Qualitative comparisons on DTU and BlendedMVS are shown in figures \ref{fig:dtu_qualitative_renders_compressed} and \ref{fig:bmvs_qualitative_renders_compressed} respectively. We picked viewpoints for which our error is highest compared to GGGS to understand the limitations of our appearance model. The texture is generally well-reproduced in our case but some specular regions are missing due to the SH bases being shared by multiple sites.

\begin{table}[!htbp]
\centering
\caption{Masked image metrics on training views.}
\label{tab:image-metrics}
\setlength{\tabcolsep}{3pt}
\begin{tabular}{lcccc}
\toprule
Dataset & Method & PSNR $\uparrow$ & SSIM $\uparrow$ & LPIPS $\downarrow$ \\
\midrule
     & CVTSplat & \thirdbest{30.32} & \best{0.888}      & \best{0.147}       \\
DTU  & GGGS     & \best{30.76}      & \secondbest{0.872}& \secondbest{0.253} \\
     & PGSR     & \secondbest{30.33}& \thirdbest{0.866} & \thirdbest{0.265}  \\
\midrule
     & CVTSplat & \thirdbest{29.52} & \thirdbest{0.894} & \secondbest{0.073} \\
BMVS & GGGS     & \best{32.43}      & \best{0.953}      & \best{0.046}       \\
     & PGSR     & \secondbest{30.50}& \secondbest{0.902}& \thirdbest{0.108}  \\
\bottomrule
\end{tabular}
\end{table}

\section{Model size} \label{sec:model_size}

Table \ref{tab:model-size} presents a comparison of our model in terms of number of primitives and number of trainable parameters, compared to PGSR~\cite{chen2024pgsr} and GGGS~\cite{Zhang2026GeometryGrounded}. Despite using more primitives than GGGS, we have a similar total parameter count on DTU, and lower on BlendedMVS. Table \ref{tab:cvtsplat-parameters} provides a breakdown of parameter usage for our approach. 

\begin{table}[!htbp]
\centering
\caption{Average model size.}
\label{tab:model-size}
\setlength{\tabcolsep}{3pt}
\begin{tabular}{lccc}
\toprule
Dataset & Method & Primitives & Parameters \\
\midrule
     & CVTSplat & 4M sites       & 39.35M \\
BMVS & GGGS     & 1.1M Gaussians & 64.21M \\
     & PGSR     & 0.5M Gaussians & 29.95M \\
\midrule
     & CVTSplat & 4M sites         & 39.35M \\
DTU  & GGGS     & 0.6M Gaussians   & 36.16M \\
     & PGSR     & 0.3M Gaussians   & 16.11M \\
\bottomrule
\end{tabular}
\end{table}

\begin{table}[!htbp]
\centering
\caption{Parameter count breakdown.}
\label{tab:cvtsplat-parameters}
\begin{tabular}{lr}
\toprule
Component & Total \\
\midrule
Site positions ($3 \times 4$M) & 12M \\
Site distances ($1 \times 4$M) & 4M \\
Site features ($4 \times 4$M)  & 16M \\
Two SH probes grids            & 4.2M \\
Environment map                & 3.1M \\
MLP                            & 595 \\
\midrule
Total                          & 39.3M \\
\bottomrule
\end{tabular}
\end{table}

\section{Details on the Multiview Loss} \label{sec:mvloss}

We use a multi-view loss inspired by PGSR \cite{chen2024pgsr}. Pairs of views that are close to each other are chosen at random during training (a mini-batch contains $4$ views so that pairs can be formed). During the forward pass of volume rendering, the depth of the first zero-crossing of the SDF is recorded for each ray to form a depth map.

\subsection{Depth Computation}

For a cell $\mathcal{C}_i$ having SDF $\hat{f}_i$, position $\mathbf{x}_i$  and SDF gradient $\mathbf{n}_i$, we have the following approximation:
\begin{align}
    \hat{f}(\mathbf{x}) = \hat{f}_i + (\mathbf{x} - \mathbf{x}_i) \cdot \mathbf{n}_i
\end{align}
Then for a ray $\mathbf{r}(t) = \mathbf{o} + t\mathbf{v}$, the zero-crossing is obtained by solving $\hat{f}(\mathbf{r}(t^*)) = 0$:
\begin{align}
    &\hat{f}_i + (\mathbf{o} + t^* \mathbf{v} - \mathbf{x}_i) \cdot \mathbf{n}_i = 0 \\
    &\Leftrightarrow \hat{f}_i + (\mathbf{o} - \mathbf{x}_i) \cdot \mathbf{n}_i + t^* \mathbf{v} \cdot \mathbf{n}_i = 0 \\
    &\Leftrightarrow t^* = - (\hat{f}_i + (\mathbf{o} - \mathbf{x}_i) \cdot \mathbf{n}_i) / \mathbf{v} \cdot \mathbf{n}_i 
\end{align}
The depth value $t^*$ is discarded if the point $\mathbf{r}(t^*)$ falls outside of the cell, indicating that the implicit surface does not cross it. There may be several zero-crossings along a given ray, so the final depth value is taken as the minimum over the intersected cells. In case no zero-crossing occurs, we use $t^* = +\infty$.

Once a depth map is obtained for both views in the pair, we can compute our multi-view loss. We implement two variants based on the zero-normalized cross-correlation (ZNCC). The first one is similar to prior works \cite{chen2024pgsr, gomez2026blobs} based on planar patches, whereas the other removes the planar assumption to recover finer details.

\subsection{Planar ZNCC}

Let the pair of views be denoted as $(A, B)$, with $A$ the reference view and $B$ the neighboring view. We compute a normal vector for each pixel $(u,v)$ in view $A$ using centered finite differences in the depth map $\mathcal{D}_A$. Let $\mathbf{n}$ be the normal vector expressed in $A$'s reference frame. Pixels having invalid depth values ($+\infty$) or invalid neighboring pixels are skipped. We check for visibility in view $B$ using a cycle consistency check (same as PGSR \cite{chen2024pgsr}):
\begin{align}
    d_A &= \mathcal{D}_A(u_A,v_A) \\
    u_B,v_B &= \pi_B(\pi_A^{-1}(u_A,v_A,d_A)) \\
    d_B &= \mathcal{D}_B(u_B,v_B) \\
    u_A',v_A' &= \pi_A(\pi_B^{-1}(u_B,v_B,d_B))
\end{align}
$\pi$ denotes the projection operator, mapping 3D points to pixel coordinates and $\pi^{-1}$ its inverse that maps a pixel and depth value back to a 3D point. We consider the pixel to be occluded in view $B$ if $\sqrt{(u_A-u_A')^2 +(v_A-v_A')^2} > 2$. The pixel may be rejected if the cycle consistency check fails due to the point projecting outside of $B$'s field of view, or if the depth in view B is invalid. Lastly, we also check that the normal is not too slanted in either view, with a rejection threshold of $85$° between the normal and the viewing ray. 

If these checks pass, we compute the ZNCC over a $5\times5$ pixel patch centered on the pixel of interest. Let $x_{i,j}$ and $y_{i,j}$ be the grayscale ground truth pixel values for views $A$ and $B$ respectively, sampled over the pixels in the patch. The grayscale values are obtained by taking the average of the RGB channels and multiplying by the mask channel. The patch forms a plane with normal $\mathbf{n}$, placed at depth $d$ along the ray originating from pixel $(u,v)$.

\begin{align}
    &x_{i,j} = \mathcal{I}_A(u+i,v+j) \qquad -2 \leq i,j \leq 2 \\
    &d' = d \frac{\mathbf{n}^T \mathbf{K}_A^{-1} (u,v,1)^T}{\mathbf{n}^T \mathbf{K}_A^{-1} (u+i,v+j,1)^T} \label{eq:ray_plane_intersection}\\
    &y_{i,j} = \mathcal{I}_B(\pi_B(\pi_A^{-1}(u+i,v+j,d')))
\end{align}

In the case of $x_{i,j}$, we can directly sample the pixels at their centers, whereas $y_{i,j}$ requires sampling image $\mathcal{I}_B$ with bilinear interpolation so that $y_{i,j}$ is a continuous function of the plane normal and depth. The sampling coordinates for $y_{i,j}$ are obtained by intersecting the plane with the neighboring rays, which gives an intersection depth $d'$, then back-projecting to 3D space and finally projecting into view B.
The ZNCC $z \in [-1,1]$ is then computed by evaluating the first- and second-order moments of the pixel samples, weighted with a Gaussian kernel.
\begin{align}
    &W = \sum w_{i,j} \label{eq:pixel_moments1}  \\ 
    &\overline{x} = \frac{1}{W} \sum x_{i,j} w_{i,j} \\ 
    &\overline{y} = \frac{1}{W} \sum y_{i,j} w_{i,j} \\ 
    &\overline{x^2} = \frac{1}{W} \sum x_{i,j}^2 w_{i,j} \\ 
    &\overline{y^2} = \frac{1}{W} \sum y_{i,j}^2 w_{i,j} \\ 
    &\overline{xy} = \frac{1}{W} \sum x_{i,j} y_{i,j} w_{i,j} \label{eq:pixel_moments2} \\ 
    &\text{ZNCC} = \frac{\overline{xy} - \overline{x} \cdot \overline{y}}{\sqrt{\overline{x^2} - \overline{x}^2 + \epsilon} \cdot \sqrt{\overline{y^2} - \overline{y}^2 + \epsilon}}
\end{align}

We take $\epsilon=1.0E{-4}$ for numerical stability. The Gaussian kernel is separable ($w_{i,j} = w_i w_j$), with the five weights taken from the 7th row of Pascal's triangle $(1, 6, 15, 20, 15, 6, 1)$, where the first and last bins are omitted since they contribute very little. 

\subsection{Nonplanar ZNCC}

Assuming that the surface is locally planar tends to erase fine geometric details so we propose to remove the planar assumption in the ZNCC computation. Doing so is simple, we just use the actual depth of the neighboring pixels in view A, instead of taking the depth of the ray/plane intersection. This amounts to replacing equation \ref{eq:ray_plane_intersection} by $d' = \mathcal{D}_A(u+i,v+j)$. In practice, we actually use a separate implementation to take advantage of the fact that the pixel values $y_{i,j}$ can be shared with neighboring pixels in view A, which is not the case in the planar case due to adjacent pixels having different plane orientations. 

We evaluate mutual visibility over each pixel in view A in a similar manner to the planar case. First, by computing the normal as before and checking its angle with respect to both viewpoints, and second, by applying the same cycle consistency check. Invalid pixels or pixels for which the mutual visibility tests fail are skipped in the computation of the moments in equations \ref{eq:pixel_moments1}-\ref{eq:pixel_moments2}.

\subsection{Loss}

Our multi-view loss is $\mathcal{L}_{\text{MV}} = (1-\text{ZNCC})$, summed over all valid pixels in view A. We use the planar ZNCC computation for most of the training since it is more stable, but switch to the nonplanar ZNCC near the end to recover finer details. Note that the grayscale ground truth images and depth maps follow the same coarse-to-fine training regime as the color losses, which means that the small $5\times5$ receptive field will actually cover a larger proportion of the field of view in the beginning. 

\subsection{Backward Pass}

The ZNCC computation performs the forward and backward passes in the same CUDA kernel for efficiency. It takes as input the depth maps $\mathcal{D}_A$ and $\mathcal{D}_B$, the grayscale images $\mathcal{I}_A$ and $\mathcal{I}_B$ and intrinsic and extrinsic parameters of both cameras. It produces the loss value $\mathcal{L}_{MV}$ along with depth gradients for view A ($\partial \mathcal{L}_{MV} / \partial \mathcal{D}_{A}$). The depth gradients are further propagated to the SDF values during the backward pass of the volume rendering.

\section{Derivations and Proofs} \label{sec:derivations_and_proofs}

This section gives more details on the computation of the SDF gradient from the neighbors' SDF (sec. \ref{proof:SDFgrad}), the computation of the volume rendering integral in closed form (\ref{proof:volume_integral}), and the check to filter non-Delaunay tetrahedra when converting the Voronoi adjacencies to a Delaunay triangulation (\ref{proof:delaunay_triangulation}). 

\subsection{SDF Gradient Computation} \label{proof:SDFgrad}

We summarize the expressions used to define the smoothed SDF $\hat{f} = G(f)$ and show derivations for its gradient $\mathbf{n} = \nabla_{\mathbf{x}} G(\hat{f})$. The smoothing kernel $G$ is defined by:
\begin{align}
    G(u)_i &= \sum_{j \in \mathcal{N}(i)} u_j w_{ij}  / \sum_{j \in \mathcal{N}(i)} w_{ij} \\
    w_{ij} &= \mathcal{V}_j \exp{\left(-\frac{||\mathbf{x}_i-\mathbf{x}_j||_2^2}{2\sigma^2 R_i^2} \right)} \\
    R_i &= \max_{j \in \mathcal{N}(i)}(||\mathbf{x}_i-\mathbf{x}_j||_2).
\end{align}
For computation of the SDF gradient, we consider that the neighborhood $\mathcal{N}(i)$, the cell volumes $\mathcal{V}_j$ and the kernel radius $R_i$ are constants. Let $W_i = \sum_{j \in \mathcal{N}(i)} w_{ij}$ be an intermediate variable to simplify the notation.
\begin{align}
    \nabla_{\mathbf{x}_i} w_{ij} &= -\frac{\mathbf{x}_i-\mathbf{x}_j}{\sigma^2 R_i^2} w_{ij} \\
    \nabla_{\mathbf{x}_i} W_i &= \sum \nabla_{\mathbf{x}_i} w_{ij} 
\end{align}
\begin{align}
    \mathbf{n}_i &= \nabla_{\mathbf{x}_i} G(\hat{f})_i \\
    &= \sum_{j \in \mathcal{N}(i)} \hat{f}_j \nabla_{\mathbf{x}_i} \left( w_{ij} / W_i \right) \\
    &= \sum_{j \in \mathcal{N}(i)} \hat{f}_j \left( \nabla_{\mathbf{x}_i} w_{ij} \cdot W_i - w_{ij} \cdot \nabla_{\mathbf{x}_i} W_i \right) / W_i^2 \\
    &= \sum_{j \in \mathcal{N}(i)} \hat{f}_j \mathbf{a}_{ij} \label{eq:sdf_grad_formula}.
\end{align}

When back-propagating loss gradients with respect to $\mathbf{n}_i$, the weights $ \mathbf{a}_{ij}$ on $\hat{f}_j$ in equation \ref{eq:sdf_grad_formula} are treated as constants.

\subsection{Closed-form Volume Integral} \label{proof:volume_integral}

Assuming the cells are not overlapping, the volume integral can be split into one integral per cell, which are then blended to produce the color of the ray $\mathbf{o} + t \mathbf{r}$. To simplify the notation, let $[a,b] = [t_i^-, t_i^+]$. For each cell, we have:
\begin{align}
    \mathbf{C}_i &= \int_{a}^{b} \mathbf{c}_i \sigma(t) T(a, t)dt \label{eq:volume_integral_cell2}\\
    T(a, t) &= e^{-\int_{a}^t \sigma(u)du} \\
    \sigma(t) &= s \Phi(-s\hat{f}(t)) |\mathbf{n}_i\cdot\mathbf{r}| \label{eq:sigma_neus_relu2}
\end{align}
Because the color $\mathbf{c}_i$ is constant inside the cell, it can be moved outside of the integral in equation \ref{eq:volume_integral_cell2}.

\begin{align}
    \int_a^b \sigma(t) T(a,t) dt 
    &= \int_a^b -\frac{d}{dt} T(a, t) dt \\
    %&= \left[- T(a,t) \right]_a^b =
    &= T(a,a) - T(a,b) = 1 - T(a,b)
\end{align}

\begin{align}
    \int_a^b \sigma(t)dt &= 
    \int_a^b s \Phi(-s\hat{f}(t)) |\mathbf{n}_i\cdot\mathbf{r}| \\
    &= s |\mathbf{n}_i\cdot\mathbf{r}| \int_a^b \Phi(-s\hat{f}(t))dt \\
    &= s |\mathbf{n}_i\cdot\mathbf{r}| \int_a^b \frac{1}{1
    + \exp(s\hat{f}(t))} dt
\end{align}

Recall that the linear SDF approximation is: 
\begin{align}
    \hat{f}(t) &= \hat{f}(\mathbf{o} + t \mathbf{r}) \\
    &= \hat{f}_i + (\mathbf{o} - \mathbf{x}_i) \cdot \mathbf{n}_i + t \mathbf{r}\cdot \mathbf{n}_i
\end{align}

Now, we perform a change of variable in the integral:

\begin{align}
    u &= s\hat{f}(t) \\
    du &= s \mathbf{r}\cdot \mathbf{n}_i dt
\end{align}

\begin{align}
\int_a^b \sigma(t)\,\mathrm{d}t
  &=
  s\left|\mathbf{n}_i \cdot \mathbf{r}\right|
  \int_{s\hat{f}(a)}^{s\hat{f}(b)}
  \frac{1}{1+e^u}\,
  \frac{\mathrm{d}u}
       {s\,\mathbf{n}_i \cdot \mathbf{r}}
  \\[4pt]
  &=
  \operatorname{sgn}\!\left(\mathbf{n}_i \cdot \mathbf{r}\right)
  \int_{s\hat{f}(a)}^{s\hat{f}(b)}
  \frac{\mathrm{d}u}{1+e^u}
  \\[4pt]
  &=
  -\operatorname{sgn}\!\left(\mathbf{n}_i \cdot \mathbf{r}\right)
  \left[
    \ln\!\left(1+e^{-u}\right)
  \right]_{s\hat{f}(a)}^{s\hat{f}(b)}
  \\[4pt]
  &=
  \operatorname{sgn}\!\left(\mathbf{n}_i \cdot \mathbf{r}\right)
  \ln\!\left(
    \frac{1+e^{-s\hat{f}(a)}}
         {1+e^{-s\hat{f}(b)}}
  \right).
\end{align}

Consequently,
\begin{align}
T(a,b)
  &= \exp\left(
       -\int_a^b \sigma(t)\,\mathrm{d}t
     \right)
  \\[4pt]
  &=
  \left(
    \frac{1+e^{-s\hat{f}(a)}}
         {1+e^{-s\hat{f}(b)}}
  \right)^{
    -\operatorname{sgn}(\mathbf{n}_i \cdot \mathbf{r})
  }.
\end{align}

This gives the final expression for the transparency:
\begin{equation}
\alpha_i = T(t_i^-,t_i^+)
=
\begin{cases}
\displaystyle
\frac{1 + \exp\!\bigl(-s\hat{f}(t_i^-)\bigr)}
     {1 + \exp\!\bigl(-s\hat{f}(t_i^+)\bigr)},
& \mathbf{n}_i \cdot \mathbf{r} < 0, \\[12pt]
\displaystyle
\frac{1 + \exp\!\bigl(-s\hat{f}(t_i^+)\bigr)}
     {1 + \exp\!\bigl(-s\hat{f}(t_i^-)\bigr)},
& \mathbf{n}_i \cdot \mathbf{r} > 0.
\end{cases}
\end{equation}

The accumulated color is then
\begin{equation}
\mathbf{C}
  = \sum_i
    \mathbf{c}_i\,(1-\alpha_i)
    \prod_{j<i} \alpha_j.
\end{equation}

\subsection{Delaunay Triangulation} \label{proof:delaunay_triangulation}

Assuming points in general position, we prove here that if a tetrahedron is not a Delaunay tetrahedron, then each of its four corners must have a Voronoi neighbor (not necessarily the same) inside its circumsphere.

By contraposition, this will show that a tetrahedron for which any of its four vertices has no Voronoi neighbors inside of the circumsphere must be a Delaunay tetrahedron.

Proof: Assume the tetrahedron IJKL is not a Delaunay tetrahedron. Without loss of generality, we look at corner I. The circumsphere of IJKL contains at least one site strictly inside of it by definition. Let C be the center of the circumsphere and P the offending site: $||C-P|| < r = ||C-I||$, with $r$ the radius of the circumsphere. In other words, C is closer to P than to I, implying that C is not inside the cell of I by definition of a Voronoi cell. 

Because C is outside of I's cell, there must be a site Q, Voronoi neighbor of I, that defines a bisector plane such that C is on the wrong side:
\begin{align}
    (C-I) \cdot (Q - I) > \frac{1}{2} ||Q - I||_2^2
\end{align}

All that remains to show is that Q must be inside the circumsphere of the tetrahedron: 
\begin{align}
    ||Q-C||^2 &= ||Q-I + I-C||^2 \\
    &= ||Q-I||^2 + 2(Q-I)\cdot(I-C) + ||I-C||^2 \\
    &= ||Q-I||^2 - \underbrace{2(Q-I)\cdot(C-I)}_{ > ||Q-I||_2^2}  + ||I-C||^2 \\
    &< ||I-C||^2 = r^2
\end{align}
We have shown that $I$ has a Voronoi neighbor $Q$ inside the circumsphere, which concludes the proof.